\documentclass[runningheads]{llncs}
\usepackage{graphicx}

\usepackage{comment}
\usepackage{amsmath,amssymb} 
\usepackage{graphicx}
\usepackage{booktabs}
\usepackage[pagebackref,breaklinks,colorlinks]{hyperref}
\usepackage{url}
\usepackage{grffile}
\usepackage{tablefootnote}
\usepackage{bm}
\usepackage[english]{babel} 
\usepackage{subcaption}
\usepackage{multirow}
\usepackage{makecell}
\usepackage{gensymb}
\usepackage{threeparttable}
\usepackage{arydshln}
\usepackage{wrapfig}
\usepackage{hhline}
\newcommand{\Tref}[1]{Table~\ref{#1}}
\newcommand{\eref}[1]{Eq.~\eqref{#1}}

\newcommand{\fref}[1]{Fig.~\ref{#1}}
\newcommand{\Fref}[1]{Figure~\ref{#1}}

\usepackage{xcolor}
\newcounter{todos}
\AtEndDocument{\ifnum\value{todos}>0 \PackageWarning{TODOS}{There are \arabic{todos} todos left in this paper! Fix them before submitting the paper!} \fi}

\newcommand{\V}[1]{\ensuremath{\mathbf{#1}}}

\usepackage{xspace}

\makeatletter
\DeclareRobustCommand\onedot{\futurelet\@let@token\@onedot}
\def\@onedot{\ifx\@let@token.\else.\null\fi\xspace}

\def\eg{\emph{e.g}\onedot} 
\def\ie{\emph{i.e}\onedot}

\def\wrt{w.r.t\onedot} 
\def\etal{\emph{et al}\onedot}
\makeatother

\begin{document}
	\pagestyle{headings}
	\mainmatter
	\def\ECCVSubNumber{4335}  
	
	\title{Edge-preserving Near-light Photometric Stereo with Neural Surfaces} 
	
	
	\titlerunning{Edge-preserving Near-light Photometric Stereo with Neural Surfaces}
	\author{Heng Guo\inst{1} \and
		Hiroaki Santo \inst{1} \and
		Boxin Shi \inst{2} \and 
		Yasuyuki Matsushita \inst{1}
	}
	\authorrunning{H. Guo et al.}
	%
	\institute{Osaka University \and
		Peking University}
\maketitle

\begin{abstract}
This paper presents a near-light photometric stereo method that faithfully preserves sharp depth edges in the 3D reconstruction.
Unlike previous methods that rely on finite differentiation for approximating depth partial derivatives and surface normals, we introduce an analytically differentiable neural surface in near-light photometric stereo for avoiding differentiation errors at sharp depth edges, where the depth is represented as a neural function of the image coordinates.
By further formulating the Lambertian albedo as a dependent variable resulting from the surface normal and depth, our method is insusceptible to inaccurate depth initialization.
Experiments on both synthetic and real-world scenes demonstrate the effectiveness of our method for detailed shape recovery with edge preservation\footnote{Source code and dataset will be made available upon acceptance.}.
\end{abstract}

\section{Introduction}

\label{sec:intro}

Photometric stereo~\cite{woodham1980ps,silver1980determining} recovers detailed surface shape from image observations under varying illuminations. While many of the photometric stereo methods assume \emph{distant} directional lights, in which light sources are placed infinitely far away from the scene, it is desirable in practice to work with a \emph{near-light} setting where light sources are placed nearby the scene.
\newcommand\imgsize{0.18}
\newcommand\imgsizew{0.2}
\begin{figure}
	\centering
	\input{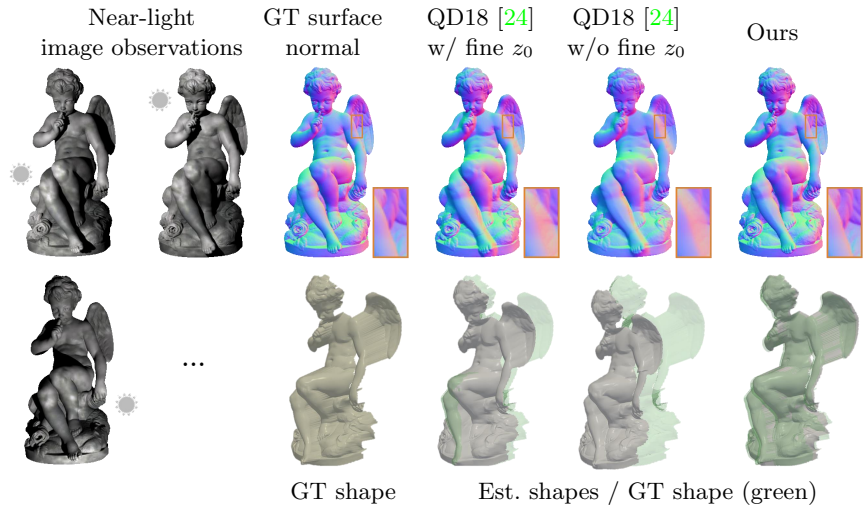}
	\caption{Given near light image observations and calibrated light positions, our method recovers accurate shapes with preserving the depth edges as highlighted in the close-up views. In addition, our method is robust against the depth initialization $z_0$ compared to the state-of-the-art methods.
	}
	\label{fig:teaser}
\end{figure}

Given near-light image observations and calibrated point light positions, near-light photometric stereo~(NLPS) aims at recovering the depth, surface normal, and reflectance at each scene point. Since a surface normal is perpendicular to the surface tangent plane modeled by depth partial derivatives, the NLPS problem naturally involves differentiation of the surface to associate it with the surface normal. 
Previous NLPS methods~\cite{queau2018led,logothetis2017semi,santo2020deep,logothetis2020cnn} represent depth and surface normal as a grid map in a discrete manner and rely on \emph{finite differentiation} to approximate the depth partial derivatives. However, the approximation does not hold well at the sharp depth edges, leading to inconsistent estimates of the surface normal and depth.

To avoid finite differentiation and achieve edge-preserving NLPS, this paper proposes the use of \emph{analytically differentiable representation} of surfaces.
As shown in \fref{fig:teaser}, the proposed method can faithfully recover the shape and surface normal, particularly better at the depth discontinuities compared to the existing state-of-the-art method~\cite{queau2018led}.
In our method, we represent the depth as an analytical function of the image coordinates, where the partial depth derivatives are modeled by the analytical expression of the function derivatives.
In this way, consistent surface normal and depth can be obtained from the analytical function without relying on finite differentiation.
Besides ensuring the depth-normal consistency, the analytical depth function is effective for representing complex and sharp depth edges. Specifically, we use a \emph{neural surface} representation based on a differentiable neural network as the analytical depth function. With the neural surface, our NLPS method is capable of detailed shape recovery with edge preservation. 

In addition, by treating the albedo as a \emph{dependent} variable resulting from the surface normal and depth, our method is made robust against inaccurate surface initialization. Compared to previous NLPS approaches that require a careful shape initialization, our method empirically exhibits preferable convergence to the accurate shape as shown in \fref{fig:teaser} (bottom) even with diverse initial shapes.

To summarize, we present an accurate NLPS method by making two technical contributions:
We propose the use of an analytical neural surface representation in NLPS for accurate shape recovery, particularly effective at the depth discontinuities. 
In addition, by treating albedos as dependent variables, we frame NLPS as an optimization of the neural surface parameters only, making our method robust against diverse initial guesses.

\section{Related Works}
\label{sec:related_work}
From the early works of Iwahori~\etal~\cite{iwahori1990reconstructing} and Clark~\cite{clark1992active}, NLPS has been continuously studied. NLPS is a non-linear problem since the surface position and light fall-off of light sources are coupled with the surface normal. Even with the Lambertian assumption and calibrated settings where the light positions are known, the problem remains difficult due to the non-linear nature of the problem.

To solve the problem, previous approaches cast the problem as a non-linear optimization~\cite{logothetis2020cnn,ahmad2014improved,bony2013tridimensional,collins20123d,huang2015near,nie2016novel,yeh2016streamlined}, in which surface normals are first estimated and depth maps are then recovered by surface normal integration techniques~\cite{xie2014surface,queau2018normal}, or use variational methods~\cite{logothetis2017semi,mecca2016single,queau2018led,mecca2014near,mecca2015realistic,liu2018near} to jointly solve for surface normals and depths via partial differential equations.
To handle non-Lambertian surfaces, NLPS methods based on the Blinn-Phong reflectance model~\cite{logothetis2017semi} and neural representation of reflectances~\cite{santo2020deep,logothetis2020cnn} have also been proposed.


In the past, the problem of depth-edge preservation in NLPS has drawn little attention despite of its importance for accurate shape recovery. 
Indeed, based on the evaluation by Mecca~\etal~\cite{mecca2021luces} on their real-world near-light photometric stereo dataset, inaccurate estimates near depth discontinuities turned out to be the main error source of existing NLPS methods. 
There has been a work that exploits the sparsity of depth discontinuities and uses $L_1$ minimization to improve the robustness of NLPS against the discontinuities~\cite{mecca2016single}. 
While their method is mathematically solid, the sparse discontinuity assumption is limited to model the real world discontinuous surfaces, such as the stairs\footnote{Results shown in the supplementary materials}, where the inaccurate finite differences along the edges cannot be treated as sparse outliers.
We consider that the error source at the depth discontinuities is due to the use of finite differentiation and propose to use an analytical surface representation, with which we can compute exact derivatives.


Another largely overlooked but important issue in NLPS is its sensitivity to the initial surface given to the solution methods. To obtain a good initial guess, Liu~\etal~\cite{liu2018near} use differential images obtained under a circular LED board. While it requires a specialized hardware, they have shown accurate shape recovery in the NLPS setting. In our method, we show that by treating albedos as dependent variables, the sensitivity to the initial guess is significantly relaxed.

\begin{figure*}
	\includegraphics[width = 1\linewidth, trim={0px 15px 0px 0px},clip]{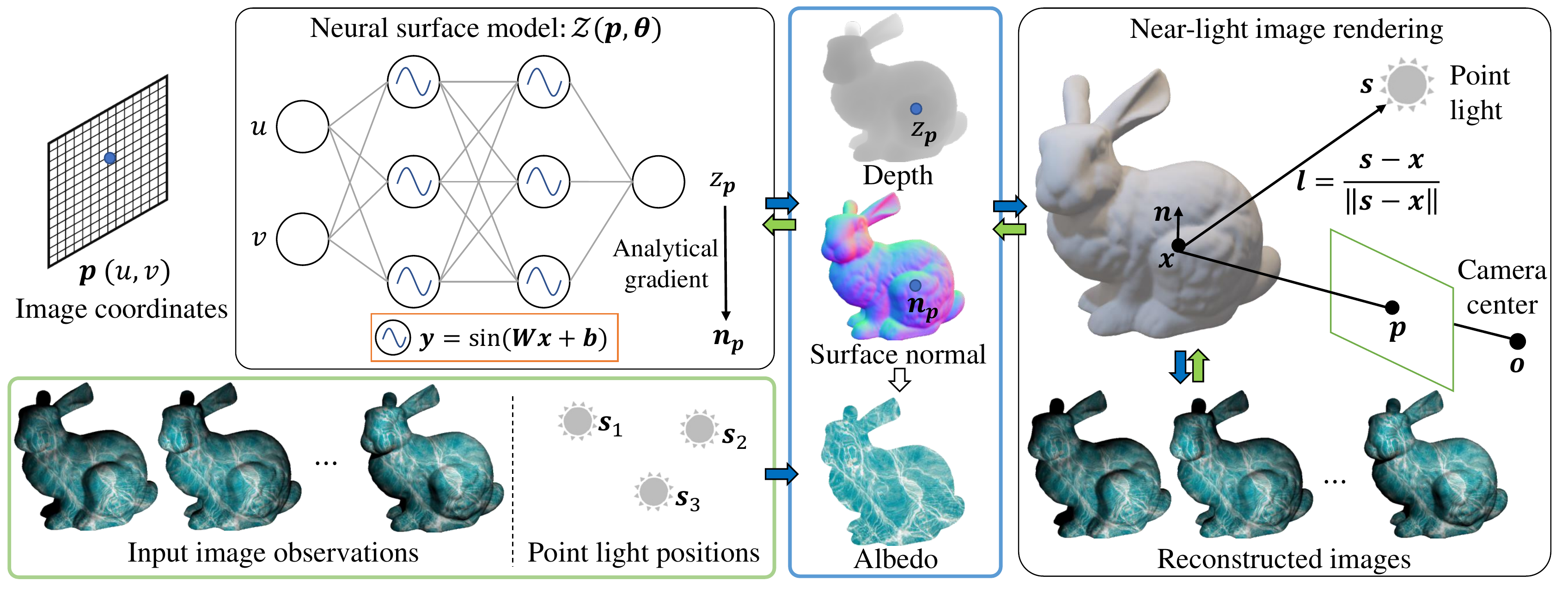}
	\caption{Pipeline of our method. The green and blue boxes show the inputs and outputs of our method, respectively.
		Following the blue arrows, we use a neural surface model that takes image coordinates as input and outputs the corresponding depth. The surface normal and albedo are derived from the depth and its analytical derivatives internally, which are used for assessing the reconstruction loss.
		Following the green arrows, we supervise the unknown neural surface parameters $\bm{\theta}$ by the difference between input image observations and the corresponding reconstructed ones.}
	\label{fig:pipeline}
	\vspace{-0.5em}
\end{figure*}

Our method is also related to analytical surface representations. In the context of shape from shading and surface normal integration, there have been works that use a Fourier basis representation~\cite{frankot1988method} and Shapelet representation~\cite{kovesi2005shapelets}.
More recently, there are deep neural network-based differentiable surface representations, such as DeepSDF~\cite{park2019deepsdf},  DISN~\cite{xu2019disn}, and LDIF~\cite{genova2020local} that are used for shape reconstruction and completion tasks. 
Sitzmann~\etal~\cite{sitzmann2020implicit} represent surface shapes with fully connected neural networks, {\sc Siren}, with periodic trigonometric functions as the activation. Their representation is differentiable and shown to have high representation power for fitting complicated signals. 
These network-based surface representations have not been applied in the context of photometric stereo so far, but we are going to show their usefulness in this paper, particularly for modeling depth discontinuities.

\section{Proposed Method} \label{sec:method}
Our method takes image observations under calibrated near-point lights as input and recovers the albedo, surface normal, and depth of the scene.
Using a neural surface model, the depth, surface normal, and albedo can all be represented by the same set of neural surface parameters $\bm{\theta}$, which are the only unknowns in our NLPS method.	
In this paper, we limit our method to assuming the Lambertian reflectance model for the purpose of demonstrating the effectiveness of the neural surface for depth edge preservation. 

The overall pipeline of the proposed method is depicted in \fref{fig:pipeline}.
In the forward pass~(blue arrows), the neural surface model takes one point in the image coordinates as input and outputs the depth and its analytical derivatives at the point simultaneously, from which we can obtain the surface normal. 
The albedo in our method is modeled as a dependent variable of the surface normal and depth. Specifically, given image observations and the shading computed from the surface normal and depth in the forward pass, the albedo is computed by minimizing the distance between the shading and image observations under the Lambertian model. 
Combining the depth, surface normal, and albedo, we reconstruct the image observations based on the near-light image formation model. In the backward pass~(green arrows), the unknown network parameters $\bm{\theta}$ are self-supervised by evaluating the reconstruction loss in a similar spirit to Taniai and Maehara's method~\cite{Taniai18}.

Throughout this paper, function \hbox{$\gamma(\cdot): \mathbb{R}^{3} \rightarrow S^{2} \; (\subset \mathbb{R}^3)$} denotes the normalization of a 3D vector, \ie, \hbox{$\gamma(\V{x}) = \frac{\V{x}}{\|\V{x}\|_2}$}.

\subsection{Near-light Image Formation Model}\label{subsec:nl_image_formation}
In NLPS, observations of a scene point is related to its albedo, surface normal, and light source positions. In this section, we first describe the geometric relationship among surface normal, depth, and scene point positions under the perspective projection, followed by the near-light image formation model. 
\paragraph{Relationship between depth and surface normal }
Denote \hbox{$\V{p} = \left[u, v\right]^\top \in \mathbb{R}^2$} as a pixel location in the image coordinates, under perspective projection, its corresponding 3D scene point \hbox{$\V{x} \in \mathbb{R}^3$} in the world coordinates is written as
\begin{eqnarray}
	\V{x} = \left[\frac{z}{f}u, \frac{z}{f}v, z \right]^\top = \left[\frac{z}{f}\V{p}^\top, z \right]^\top,
	\label{eq:x_repre_pers}
\end{eqnarray}
where \hbox{$z \in \mathbb{R}$} is the depth value at \hbox{$\V{p}$}, and $f$ is the camera's focal length. 

Surface normal \hbox{$\V{n} \in S^{2} \subset \mathbb{R}^3$} is a unit vector that is perpendicular to the tangent plane at $\V{x}$. 
Following Qu{\'e}au~\etal\cite{queau2018normal}, the surface normal under perspective projections is written as
\begin{eqnarray}
	\V{n} = \gamma \left( \begin{bmatrix}
		f\nabla z,\\
		-z - \nabla z^\top\V{p}
	\end{bmatrix}\right),
	\label{eq:n_repre_pers}
\end{eqnarray}
where \hbox{$\nabla z = \left[\frac{\partial z}{\partial u}, \frac{\partial z}{\partial v}\right]^\top$} is the partial derivatives of the depth.
Our method can work with both orthogonal and perspective projections. From a practical viewpoint, we use the perspective camera model in the rest of the paper.

\paragraph{Near-light image formation}
Under the near-light setting, the light direction at a scene point $\V{x} \in \mathbb{R}^3$ is given by
\begin{eqnarray}
	\V{l} = \gamma(\V{q} - \V{x}),
\end{eqnarray}
where $\V{q} \in \mathbb{R}^3$ is the light position. It contributes the shading $s \in \mathbb{R}_+$ on $\V{p}$ as
\begin{eqnarray}
	s = \Phi \frac{1}{\|\V{q} - \V{x} \|_2^2} {\rm max}(\V{l}^\top \V{n}, 0),
\end{eqnarray}
where \hbox{$\rm max(\cdot, 0)$} accounts for attached shadows, $\frac{1}{\|\V{q} - \V{x} \|_2^2}$ models the light fall-off, $\Phi \in \mathbb{R}_+$ is the radiant intensity of the light source.
For anisotropic light sources, the radiant intensity is modeled by~\cite{mecca2016single,queau2018led}
\begin{eqnarray}
	\Phi = \Phi_0 (\V{l}^\top \bm{\omega})^\mu,
	\label{eq:phi_repre}
\end{eqnarray}
where the radiant parameters $\Phi_0$, $\mu$, and $\bm{\omega}$ are the light irradiance, principal direction and the attenuation parameters. Similar to previous works~\cite{queau2018led,park2014calibrating}, we assume both light source positions and radiant parameters are known from light calibration. 
Combining Eqs.~(\ref{eq:x_repre_pers}) to (\ref{eq:phi_repre}) and removing the redundant variables including $\V{l}, \V{n}$, and $\V{x}$, the shading $s$ becomes a function $\mathcal{S}$ of pixel position $\V{p}$, point light source position $\V{q}$, depth $z$, and its derivatives $\nabla z$, \ie,
\begin{eqnarray}
	s = \mathcal{S}(\V{p}, \V{q}, z, \nabla z),
	\label{eq:shading_func}
\end{eqnarray}
where the radiant parameters ($\Phi_0$, $\mu$, and $\bm{\omega}$) and camera's focal length $f$ are constant and thus omitted.

If the global illumination effects, \eg, cast shadows and inter-reflections, are negligible, and the camera has a linear response, then the Lambertian image observation at pixel $\V{p}$ under the near point light at $\V{q}$ can be written as
\begin{eqnarray}
	m_{\V{p},\V{q}} = \rho_\V{p}\mathcal{S}\left(\V{p}, \V{q}, z_\V{p}, \nabla z_\V{p}\right),
	\label{eq:nl_rendering}
\end{eqnarray}
where $\rho_\V{p} \in \mathbb{R}_{+}$ is the diffuse albedo. 
Given image observations and calibrated light source positions, NLPS aims at recovering depth and albedo via minimizing
\begin{eqnarray}
	\label{eq:energy}
	\begin{aligned}
		(\V{z}^*, &\bm{\rho}^*)= 
		\operatornamewithlimits{argmin}\limits_{\V{z}, \bm{\rho}} \sum_\V{p, q} \left( m_{\V{p}, \V{q}} -  \rho_\V{p}\mathcal{S}(\V{p}, \V{q}, z_\V{p}, \nabla z_\V{p})\right)^2.
	\end{aligned}
\end{eqnarray}

Our NLPS method uses a similar energy function to previous methods but represents both depth and albedo with neural surface parameters, as we will show in the following subsection.

\subsection{Edge-preserving Near-light Photometric Stereo}\label{subsec:nlps}
\begin{figure}
	\resizebox{1\textwidth}{!}{
	\centering
	\begin{threeparttable}
		\begin{tabular}{ccc}
			\toprule
			\makecell[c]{\textbf{GT} \\ \textbf{surface}}
			\makecell[l]{
				\parbox[l]{0.3\textwidth}{\begin{equation*}
						\mathcal{Z}(u, v) = \begin{cases}
							1 & u>0, v>0, \\
							0 & {\rm Others}
						\end{cases}
			\end{equation*}}}
			& \makecell[c]{\textbf{Discrete surface} \\ \textbf{representation}}  
			\makecell[l]{
				\parbox[l]{0.3\textwidth}{\begin{equation*}
						z_\V{p} = \begin{cases}
							1 & \V{p} \in \Omega_1, \\
							0 & \V{p} \in \Omega_0 
						\end{cases}
			\end{equation*}}}
			& \makecell[c]{\textbf{Analytical surface} \\ \textbf{representation}} 
						\makecell[l]{
				\parbox[l]{0.3\textwidth}{\begin{equation*}
						\mathcal{Z}(u, v; k) = \sigma(k u) \cdot \sigma(k v) 
			\end{equation*}}}
			\\
			\midrule
			\includegraphics[width=0.5\textwidth]{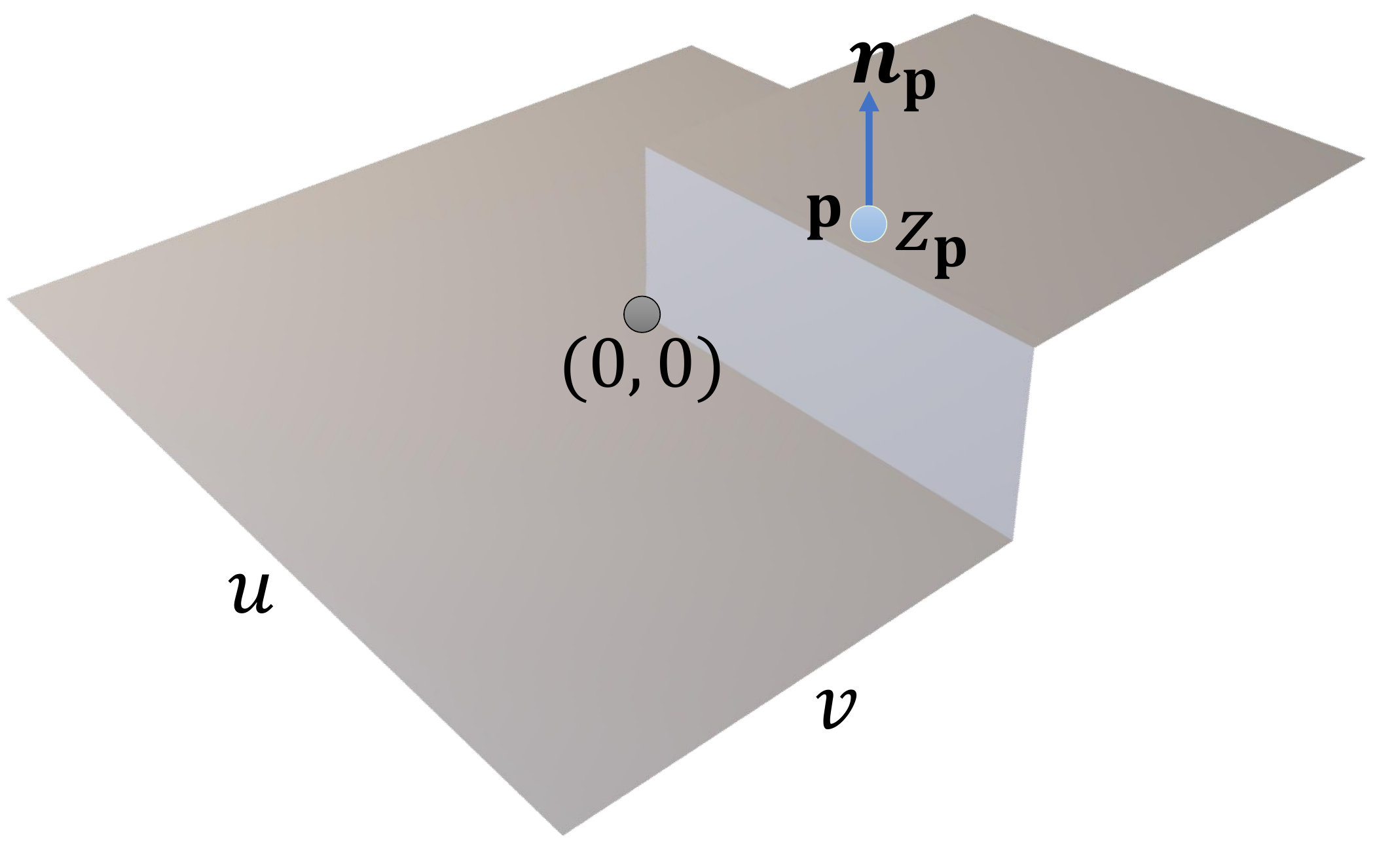}
			& \includegraphics[width=0.5\textwidth]{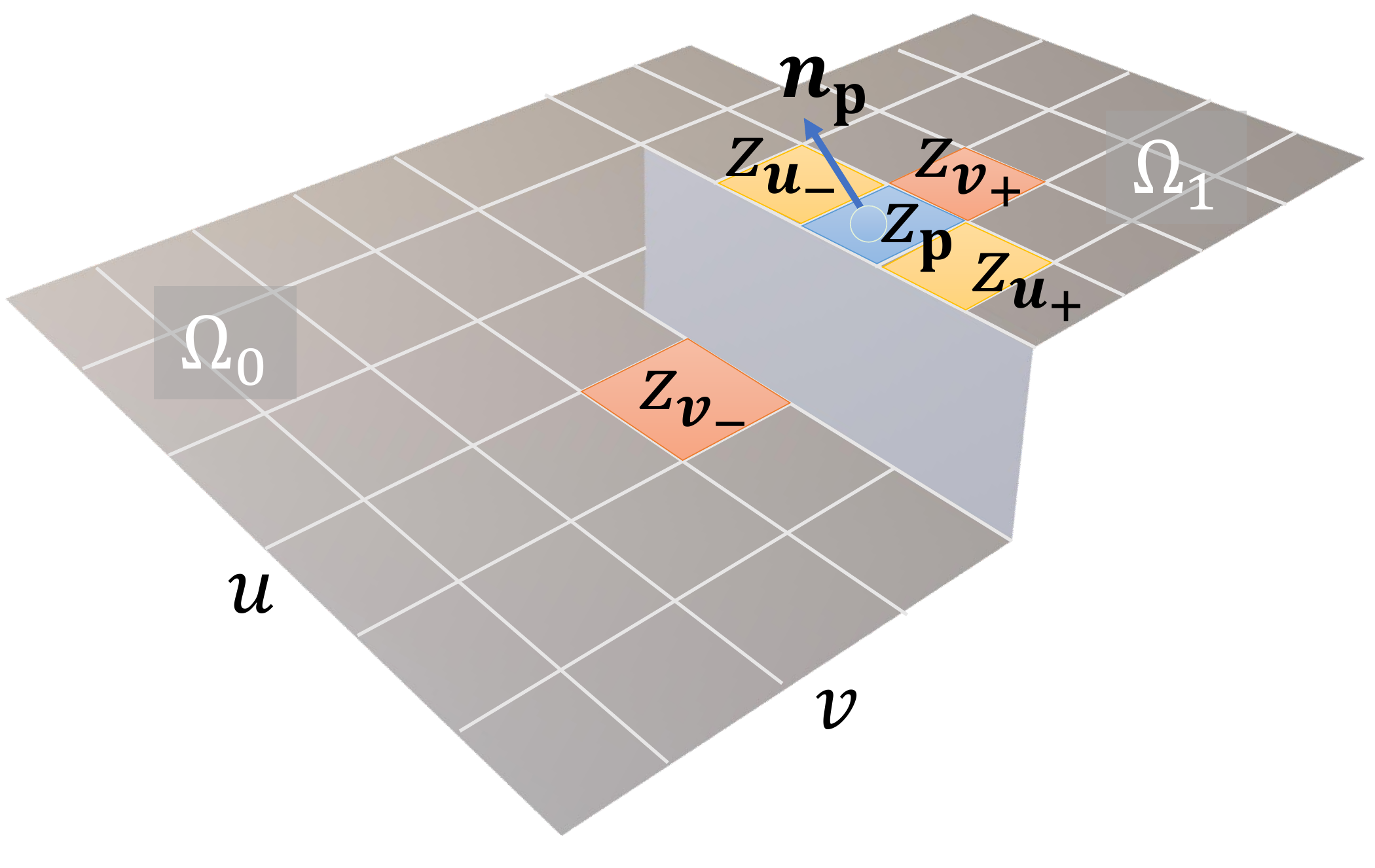}
			& \includegraphics[width=0.5\textwidth]{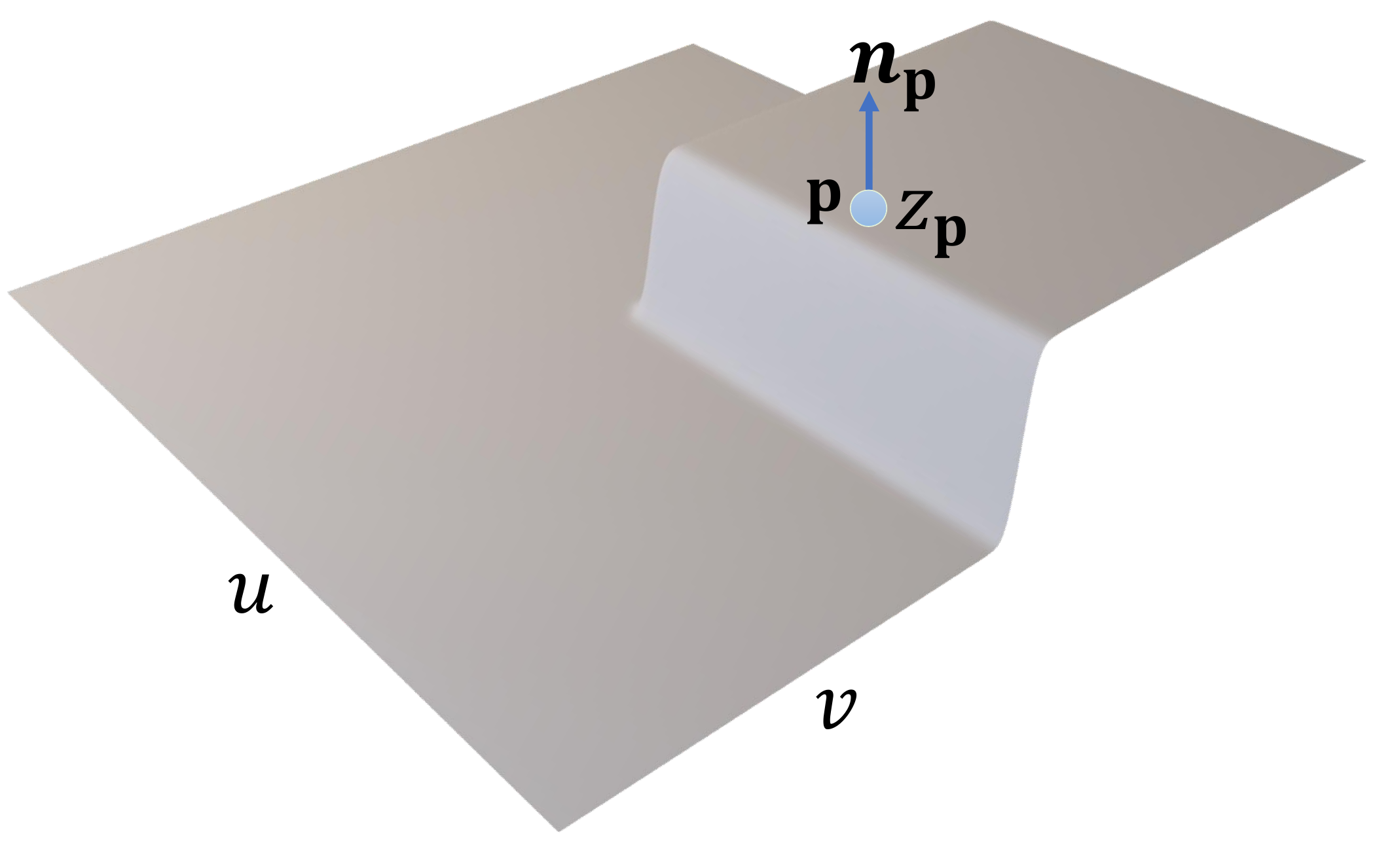} 
			\\
			GT derivative
			& 
			Finite difference approximation
			&
			Analytical derivative
			\\
			\makecell[l]{
						\parbox[l]{0.5\textwidth}{\begin{equation*}
					\nabla z_{\V{p}} =  \begin{bmatrix}
						\frac{\partial z_\V{p}}{ \partial u}, 
						\frac{\partial z_\V{p}}{ \partial v}
					\end{bmatrix}^\top
					=  \begin{bmatrix}
						0 \\
						0
					\end{bmatrix}
			\end{equation*}}}
			&
			\makecell[l]{
			\parbox[l]{0.5\textwidth}{\begin{equation*}
				 \nabla z_{\V{p}} \approx  \begin{bmatrix}
				 	 \frac{z_{\V{u}+} - z_{\V{u}-}}{ 2 \Delta u},
 				 	 \frac{z_{\V{v}+} - z_{\V{v}-}}{ 2 \Delta v}
				 \end{bmatrix}^\top
			 =  \begin{bmatrix}
			 	0 \\
			 	0.5
			 \end{bmatrix}
			\end{equation*}}}
			&
			\makecell[l]{
				\parbox[l]{0.5\textwidth}{\begin{equation*}
						\nabla z_{\V{p}}
						= \begin{bmatrix}
							k \sigma(k u)\sigma(k v) (1 - \sigma(k u)) \\
							k \sigma(k u)\sigma(k v) (1 - \sigma(k v))
						\end{bmatrix}
						\approx \begin{bmatrix}
							0 \\
							0
						\end{bmatrix}
			\end{equation*}}}
		\\
		\bottomrule
		\end{tabular}
	\end{threeparttable}	
}

	\caption{Illustration of finite and analytical differentiation at a near-edge scene point $\V{p}$ shown in blue. The finite differentiation results in inaccurate derivatives $\nabla z_\V{p}$ and surface normals $\V{n}_\V{p}$ along depth edges. Analytical surface representation~(\eg, product of sigmoid functions $\sigma(\cdot)$) provides consistent analytical derivatives and surface normals to the depth, which could be more accurate with a proper setting of parameter $k$.}
	\vspace{-1em}
	\label{fig:motivation}
\end{figure}

In this section, we introduce our NLPS with a neural surface model to handle sharp edges and solve the neural surface parameters with the reconstruction loss.

We begin with depicting the distinction between finite and analytical surface differentiations.
\Fref{fig:motivation} shows a simple stair surface represented in a discrete and an analytical manner, respectively. The depth derivatives for a near-edge scene point $\V{p}$ are approximated inaccurately via the finite difference in the discrete surface model, leading to inconsistent depth and surface normal. With analytical surface representation, where the stair surface is represented by the product of two sigmoid functions $\sigma(\cdot)$ in this particular example, 
the surface normal and depth are guaranteed to be consistent for every scene point from the analytical expression, and there is no approximation in computing the analytical depth derivatives.

Our neural surface belongs to analytical surface representation and is empowered by flexibility of neural networks to model sharp edges. We represent the surface depth as an analytical neural function of 2D image coordinates $\mathcal{Z}$, \ie, \hbox{$z_\V{p} = \mathcal{Z}(\V{p}; \bm{\theta})$}, where $\bm{\theta}$ encodes all the network parameters.
As the analytical neural surface is differentiable \wrt the pixel position, the first order partial derivatives of the depth $\nabla z_\V{p}$ can be exactly calculated as
\begin{eqnarray}
	\nabla z_\V{p} = \begin{pmatrix}
		\mathcal{Z}_u(\V{p}; \bm{\theta})\\
		\mathcal{Z}_v(\V{p}; \bm{\theta})
	\end{pmatrix} = \nabla \mathcal{Z}(\V{p}; \bm{\theta}),
	\label{eq:par_derivative}
\end{eqnarray}
which is also an analytical function of the neural surface parameters $\bm{\theta}$. Therefore, the surface normal from the depth partial derivatives can be obtained together with the depth (see \eref{eq:n_repre_pers}) from the same set of parameters. 
If the neural surface is sufficiently flexible to represent the true depth, both the surface normal and depth can be accurately represented by the network parameters, including the areas with sharp edges and high-frequency curvatures.

%

Theoretically, any analytical surface models can be used in our NLPS method. In this paper, we use {\sc Siren}~\cite{sitzmann2020implicit} as our neural surface model, which has a strength in modeling finely detailed signals.
Taking a single {\sc Siren} neuron as an example, the depth is represented by a weighted sum of a set of sine functions with varying frequencies as
\begin{eqnarray}
	\mathcal{Z}(\V{p}; \bm{\theta}) = \V{w}_1^\top\sin(\V{W}_2\V{p} + \V{b}_2) + b_1,
	\label{eq:siren}
\end{eqnarray} 
where $\V{w}_1 \in \mathbb{R}^{k}$ and $b_1 \in \mathbb{R}$ are the coefficients and bias of the sine functions, respectively, and $\V{W}_2 \in \mathbb{R}^{k \times 2}$ and $\V{b}_2 \in \mathbb{R}^{k}$ encode the $k$ diverse frequencies and phases in the sine functions. All these parameters are embedded in $\bm{\theta}$. 
With more neurons in the network, we can represent high-frequency curvatures such as sharp edges. As the {\sc Siren} network is differentiable, we can obtain the analytical depth derivatives as
\begin{eqnarray}
	\nabla \mathcal{Z}(\V{p}; \bm{\theta}) = \V{W}_2^\top{\rm diag}(\cos(\V{W}_2\V{p} + \V{b}_2))\V{w}_1,
	\label{eq:siren_derivative}
\end{eqnarray} 
where ${\rm diag(\cdot)}$ is a diagonalization operator. Based on the analytical derivatives, the surface normal vector can be obtained following \eref{eq:n_repre_pers}.

\paragraph{Reconstruction loss}
We use the near-light image formation model to optimize the neural surface parameters $\bm{\theta}$ by evaluating the reconstruction loss.
Our method treats albedos that appear in the image formation model as dependent variables of surface normals and depths. Thus, the albedos are also governed by the neural surface parameters $\bm{\theta}$.

By rewriting the Lambertian near-light image formation model of \eref{eq:nl_rendering} with neural surface parameters $\bm{\theta}$, we have
\begin{eqnarray}
	\begin{aligned}
		m_{\V{p},\V{q}}
		&= \rho_\V{p} \mathcal{T}(\V{p}, \V{q}, \mathcal{Z}(\V{p}; \bm{\theta}), \nabla \mathcal{Z}(\V{p}; \bm{\theta}))\\
		&\equiv \rho_\V{p} \mathcal{T}(\V{p}, \V{q}; \bm{\theta}).
	\end{aligned}
	\label{eq:img_formation_siren}
\end{eqnarray}
Under the illumination of $f$ near-point lights, we have a vector of image observations \hbox{$\V{m}_\V{p} = \left[m_{\V{p}, \V{q}_1}, \cdots, m_{\V{p}, \V{q}_f}  \right]^\top$}. The corresponding shading vector is obtained by \hbox{$\V{s}_\V{p} = \left[\mathcal{S}(\V{p}, \V{q}_1; \bm{\theta}), \cdots, \mathcal{S}(\V{p}, \V{q}_f; \bm{\theta})\right]^\top$}.
From the measurement and shading vectors, we obtain the least-squares approximate albedo $\hat{\rho}_\V{p}$ as
\begin{eqnarray}
	\hat{\rho}_\V{p} = \frac{\V{m}_\V{p}^\top \V{s}_\V{p}}{\V{s}_\V{p}^\top \V{s}_\V{p}}.
	\label{eq:albedo_lsqr}
\end{eqnarray}
Because the shading $\V{s}_{\V{p}}$ is a function of the neural surface parameters $\bm{\theta}$ and the image observations $\V{m}_\V{p}$ are constant, the albedo only depends on the network parameters $\bm{\theta}$. In this manner, the NLPS optimization shown in \eref{eq:energy} reduces from depth and albedo to the neural surface parameters $\bm{\theta}$ only.

Based on the the least-squares approximate albedo, surface normal, and depth, we reconstruct the images based on \eref{eq:img_formation_siren}. The reconstruction loss is evaluated by the $L_1$ distance between input image observations and the reconstructed ones as
\begin{eqnarray}
	\mathcal{L} = \frac{1}{pf}\sum_\V{p, q} \left| m_{\V{p}, \V{q}} -  \hat{\rho}_\V{p} \mathcal{S}(\V{s}, \V{p}; \bm{\theta}) \right|,
	\label{eq:loss_ours}
\end{eqnarray} 
where $p$ is the number of pixels. 
Supervised by the reconstruction loss, we update the neural surface parameters $\bm{\theta}$ in the backward process until convergence. From optimized neural surface parameters, the surface normal, depth, and albedo can be reconstructed accordingly. Empirically, our loss based on the albedos as dependent variables exhibits robustness against the initial guess, as we will see in the experiment section. 

\paragraph{Implementation}
Our neural surface model consists of $5$ fully-connected hidden layers with a sine activation function and $1$ linear output layer. Each layer contains $256$ channels. 
Overall, this $6$-layer network is rather lightweight with $0.3$M learnable parameters compared to the non-parametric depth map representation~\cite{santo2020deep}. 
We initialize the network parameters $\bm{\theta}$ following the setting of {\sc Siren}~\cite{sitzmann2020implicit}. The image coordinates in the network are scaled to the range of $[-1, 1]$. 
We adopt the Adam optimizer~\cite{kingma2014adam} with default settings ($\beta_1 = 0.9$ and $\beta_2 = 0.99$). The learning rate is initially set to $1 \times 10^{-4}$ and divided by $2$ at every $8000$ iterations. The optimization is terminated when the reconstruction loss becomes lower than $1 \times 10^{-6}$.
Besides, to ignore the influence of cast shadows, we discard shadowed pixels based on a thresholding strategy similar to~\cite{shi2019benchmark}, \ie, discard the pixels with intensities less than $5\%$ of the median of the intensities in each image.
Please refer to our supplementary material for more details.

\renewcommand\imgsize{0.2}
\renewcommand\imgsizew{0.2}
\begin{figure*}
	\centering
	\resizebox{0.75\textwidth}{!}{
		\begin{tabular}{ccccc}
			{\sc Tent} & {\sc Bunny} & {\sc Buddha} & {\sc Cupid} & {\sc Bear} \\
			{\includegraphics[ height=\imgsize\textwidth, trim={65px 65px 65px 65px},clip]{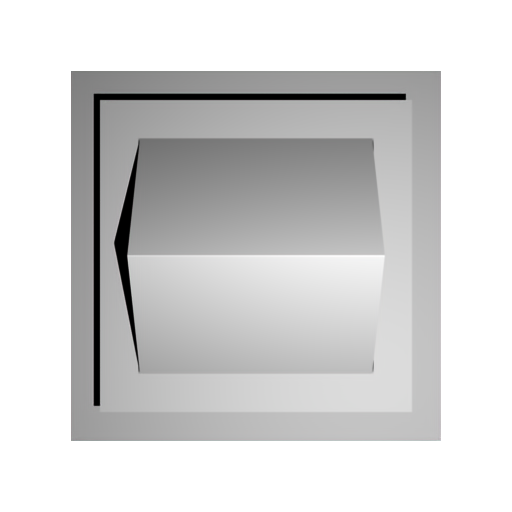}}
			&{\includegraphics[ height=\imgsize\textwidth, trim={10px 0px 40px 33px},clip]{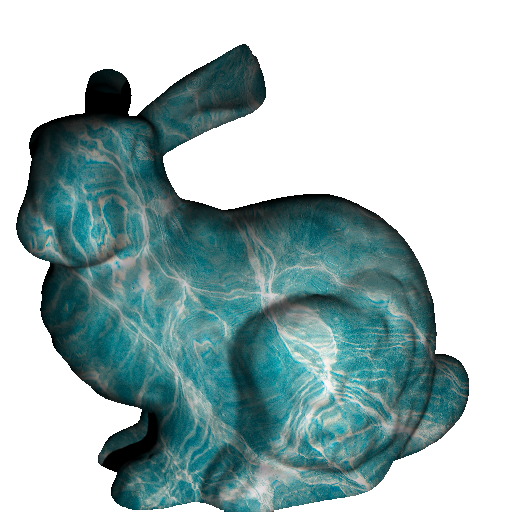}}
			&{\includegraphics[ height=\imgsize\textwidth, trim={112px 23px 112px 40px},clip]{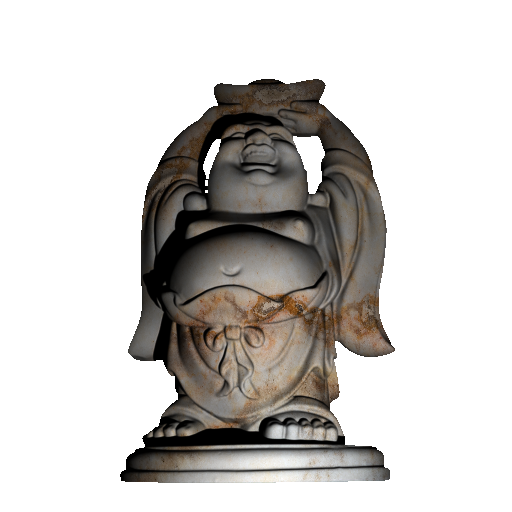}}
			&{\includegraphics[ height=\imgsize\textwidth, trim={112px 23px 112px 40px},clip]{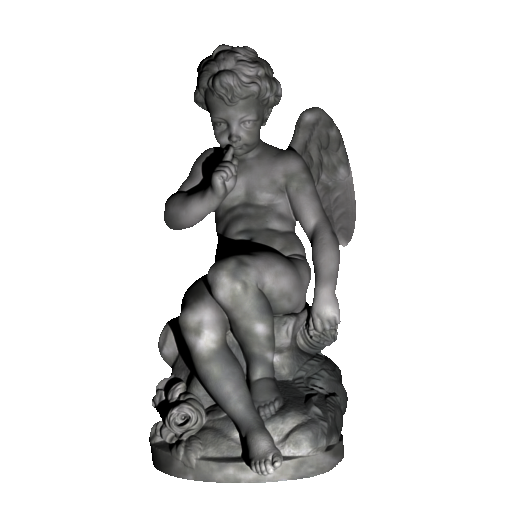}}
			&{\includegraphics[ height=\imgsize\textwidth, trim={0px 0px 0px 0px},clip]{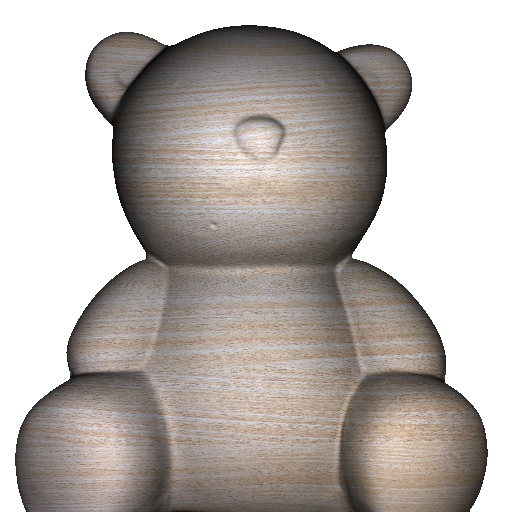}}
		\end{tabular}	
}
	\caption{Synthetic near-light image dataset in our experiment.}
		\vspace{-1em}
	\label{fig:synthetic_data}
\end{figure*}

\section{Experiments}
We evaluate our method on both synthetic and real-world datasets. We choose QD18~\cite{mecca2016single} and LB20~\cite{logothetis2020cnn} as the baseline methods, which are the state-of-the-art NLPS methods for Lambertian and non-Lambertian reflectances.
Additional comparisons with MQ16~\cite{mecca2016single} and SM20~\cite{santo2020deep} are summarized in the supplementary material.
Following the initialization setting of the baseline methods, we use a plane located at the mean distance between the object and camera as the initial shape for the existing method.

\subsection{Experiments on Synthetic Dataset}
We render five synthetic objects shown in \fref{fig:synthetic_data} with a Lambertian reflectance using the renderer Blender 2.80~\cite{blendercite}.
The camera is placed at the world origin with focal length and image resolution of $50$~mm and $512 \times 512$, respectively. To synthesize realistic appearances, the renderings include cast shadows.
For the lighting condition, we use $81$ point light sources arranged in a $9 \times 9$ regular grid with a uniform intensity~($\mu = 0, \Phi_0 = 1$). The light source locations are in the range of $(X, Y, Z) = (\pm 1, \pm 1, 0)$, with the unit of meter. The object is scaled and roughly located in the range of $(X, Y, Z) = (\pm 1, \pm 1, 3 \pm 0.5)$.

\begin{figure*}
		\resizebox{0.95\textwidth}{!}{
		\large
		\begin{tabular}{@{}c:c@{}@{}c@{}@{}c:c@{}@{}c@{}@{}c@{}}
			GT shape
			&\multicolumn{3}{c}{3D shape estimation}   &\multicolumn{3}{:c}{Depth absolute error} \\
			{\includegraphics[ width=\imgsize\textwidth, trim={44px 56px 38px 88px},clip]{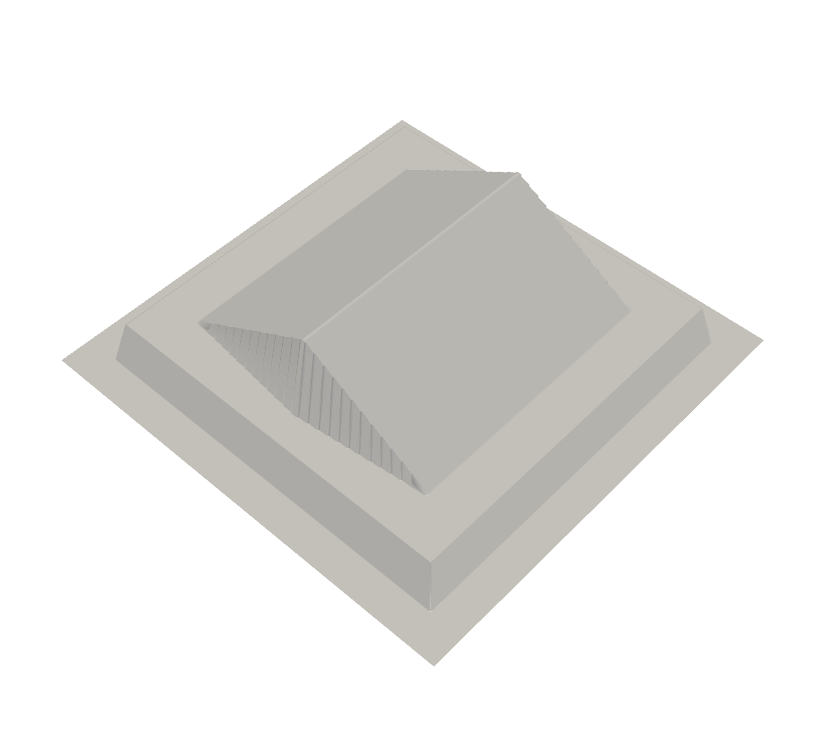}}
			&{\includegraphics[ width=\imgsize\textwidth, trim={44px 56px 38px 88px},clip]{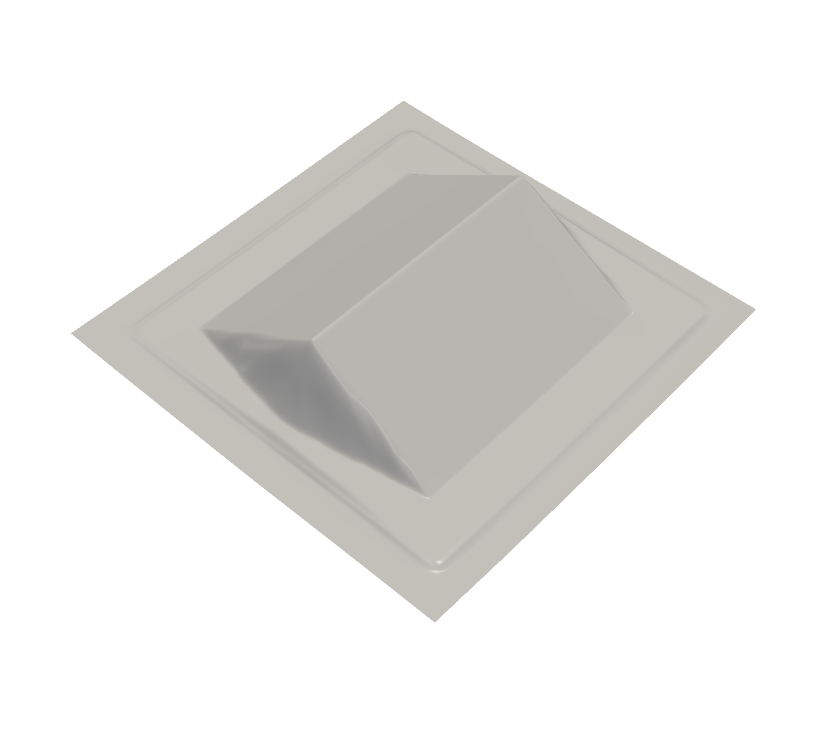}}
			&{\includegraphics[ width=\imgsize\textwidth, trim={44px 56px 38px 88px},clip]{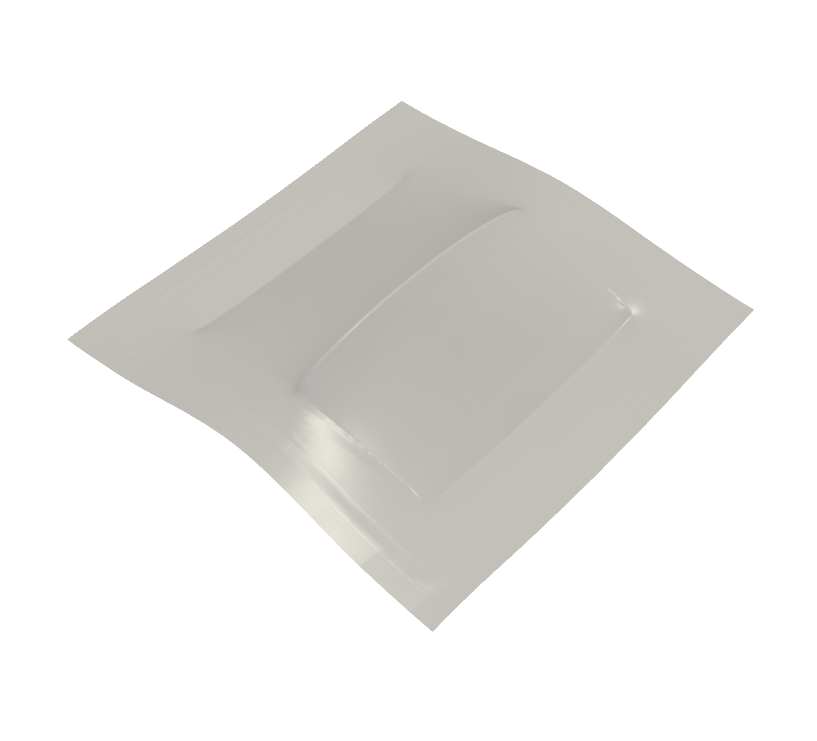}}
			&{\includegraphics[ width=\imgsize\textwidth, trim={44px 56px 38px 88px},clip]{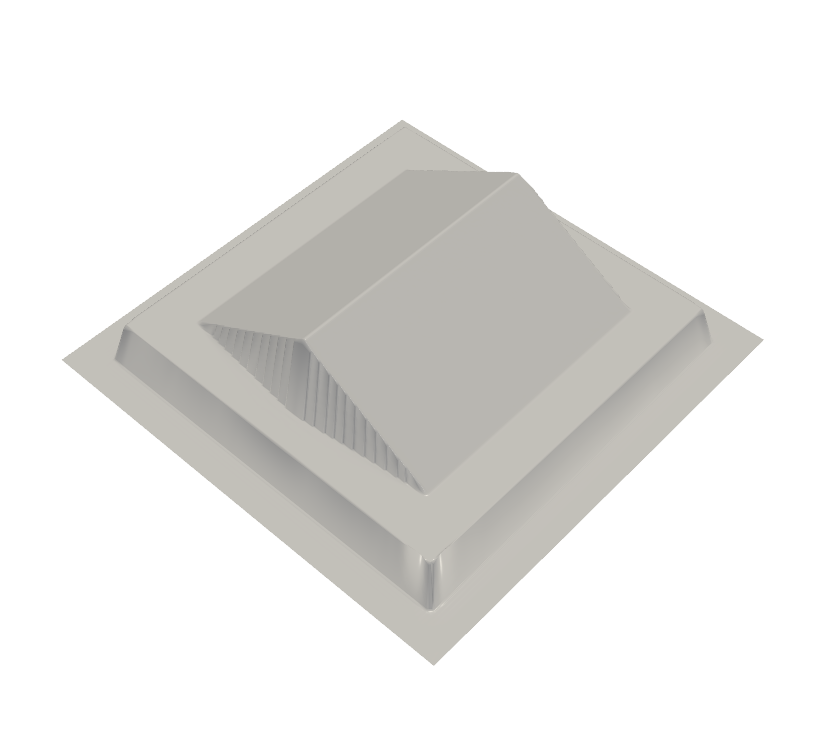}}
			&{\includegraphics[ width=\imgsize\textwidth, trim={65px 65px 65px 65px},clip]{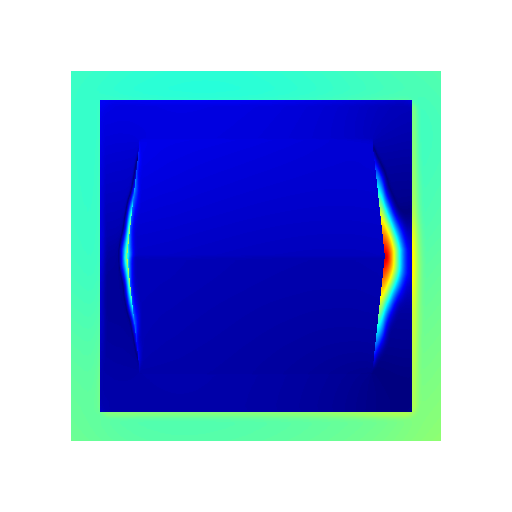}}
			&{\includegraphics[ width=\imgsize\textwidth, trim={65px 65px 65px 65px},clip]{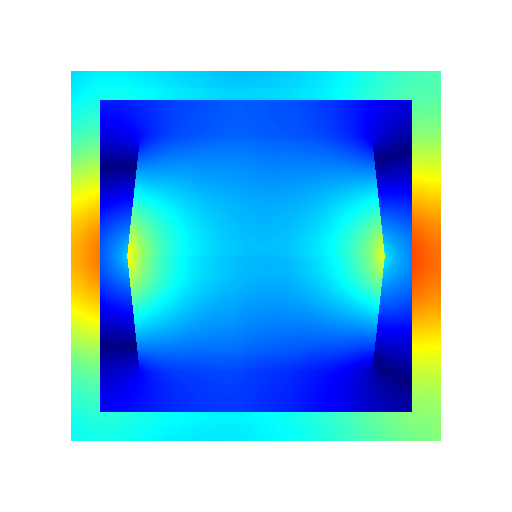}}
			&{\includegraphics[ width=\imgsize\textwidth, trim={65px 65px 65px 65px},clip]{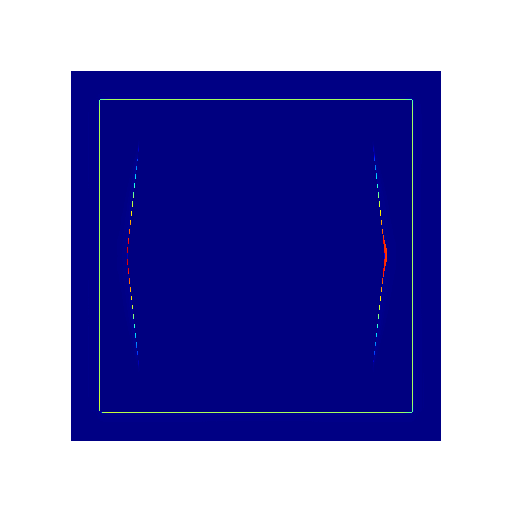}}\\
			{\includegraphics[ width=\imgsize\textwidth, trim={145px 93px 184px 119px},clip]{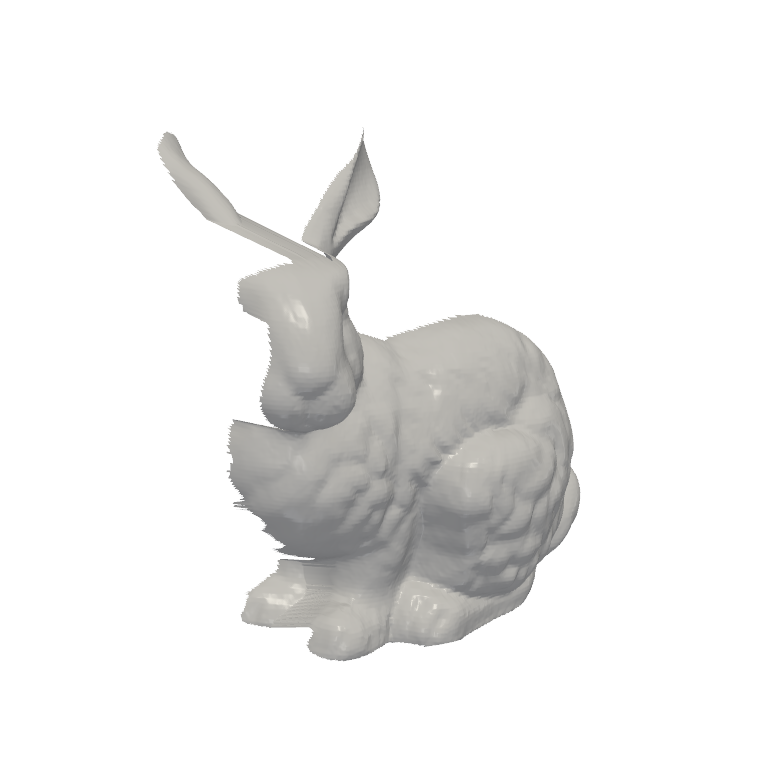}}
			&{\includegraphics[ width=\imgsize\textwidth, trim={145px 93px 184px 119px},clip]{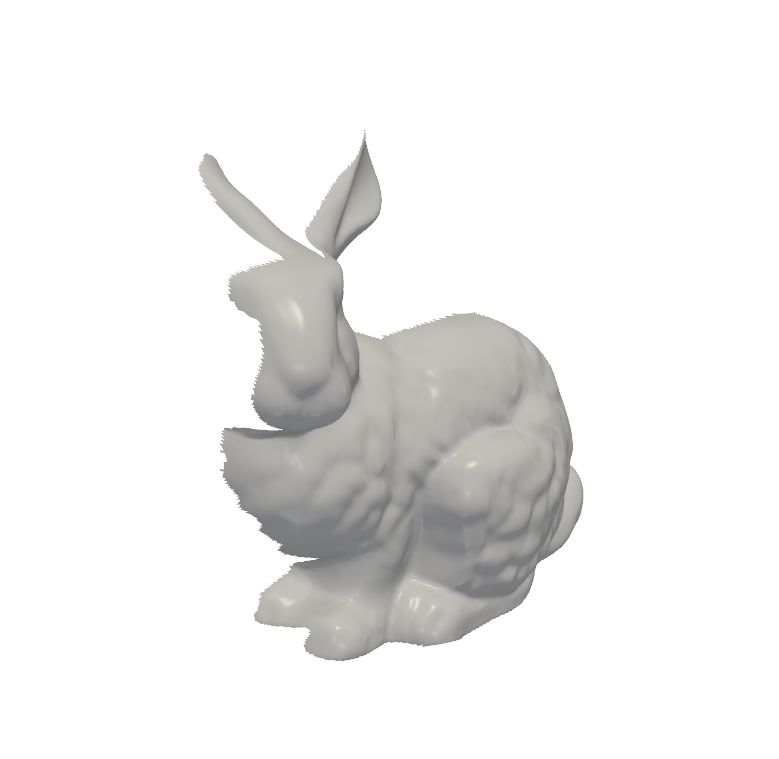}}
			&{\includegraphics[ width=\imgsize\textwidth, trim={145px 93px 184px 119px},clip]{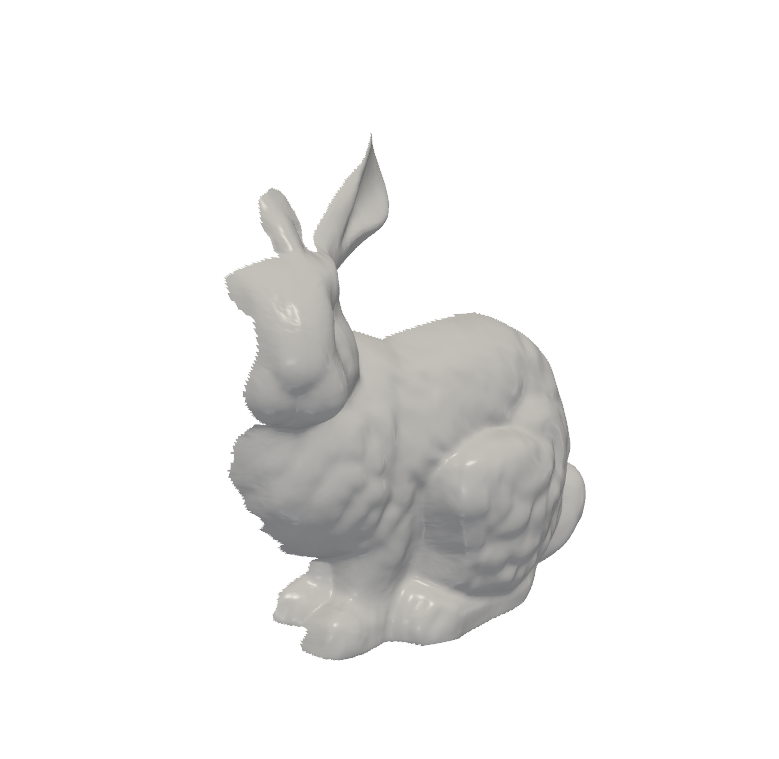}}
			&{\includegraphics[ width=\imgsize\textwidth, trim={145px 93px 184px 119px},clip]{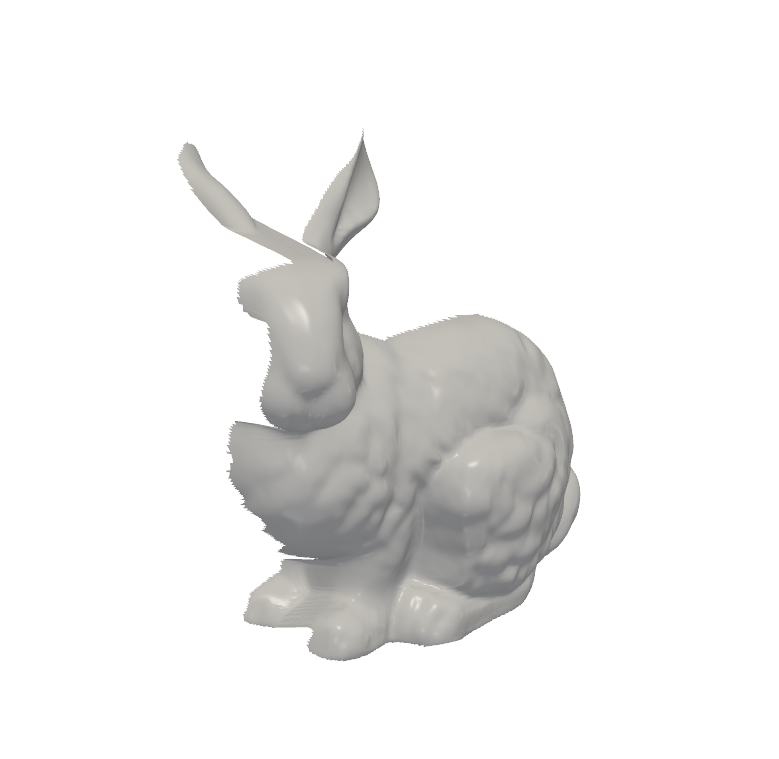}}
			&{\includegraphics[ width=\imgsize\textwidth, trim={10px 0px 40px 33px},clip]{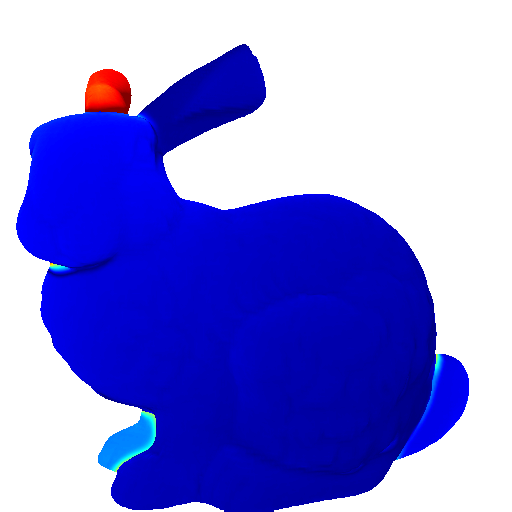}}
			&{\includegraphics[ width=\imgsize\textwidth, trim={10px 0px 40px 33px},clip]{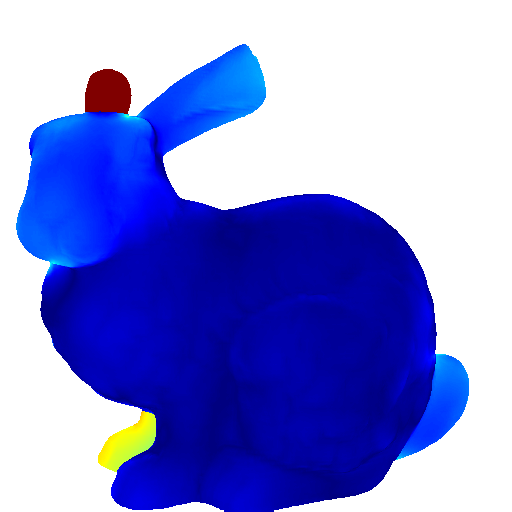}}
			&{\includegraphics[ width=\imgsize\textwidth, trim={10px 0px 40px 33px},clip]{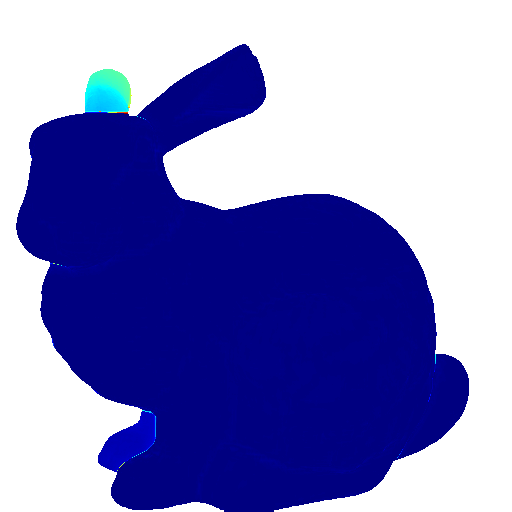}} 
			\begin{minipage}{0.013\textwidth} \centering
				\vspace{-5.5em} \makebox[0.5\textwidth]{ $ 0.1\,{\rm m}$}\\ \vspace{-0em}
				\makebox[0.5\textwidth]{}\includegraphics[width=0.9\linewidth]{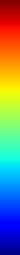} \\ \vspace{-0.4em}
				\makebox[0.5\textwidth]{ $0\,{\rm m}$}\\
			\end{minipage}\\
			{\includegraphics[ width=\imgsize\textwidth, trim={107px 47px 150px 43px},clip]{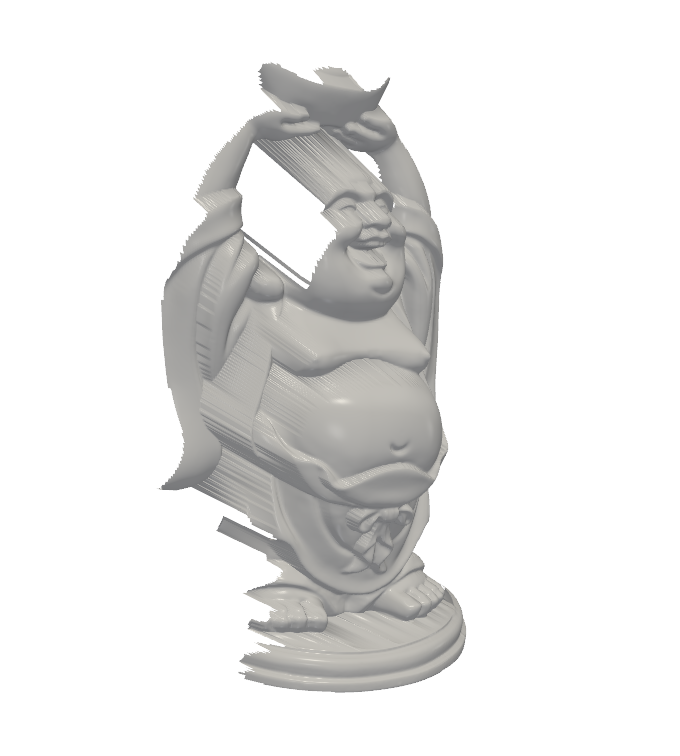}}
			&{\includegraphics[ width=\imgsize\textwidth, trim={107px 47px 150px 43px},clip]{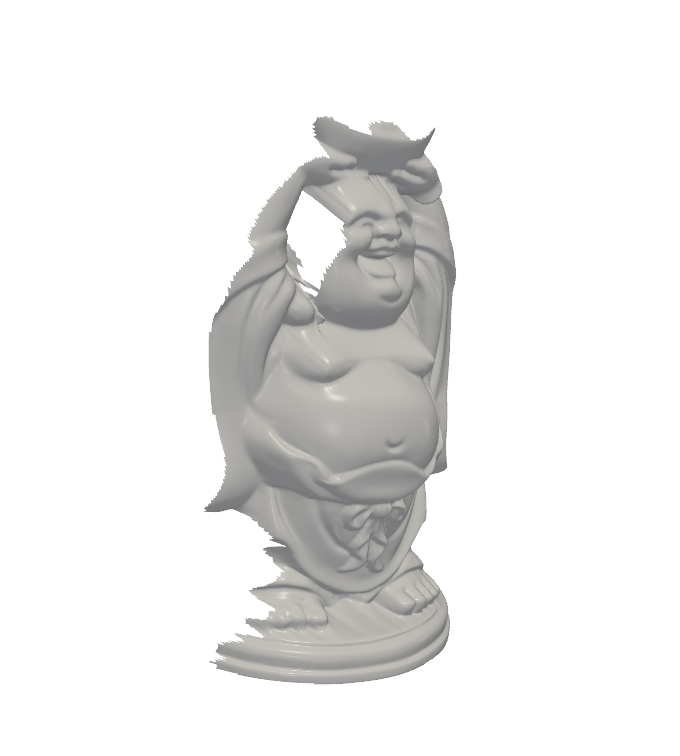}}
			&{\includegraphics[ width=\imgsize\textwidth, trim={107px 47px 150px 43px},clip]{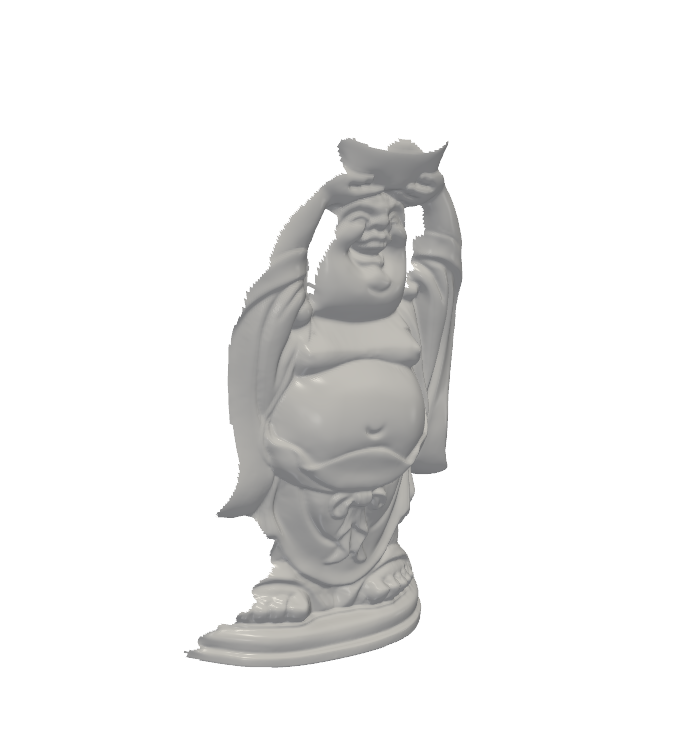}}
			&{\includegraphics[ width=\imgsize\textwidth, trim={107px 47px 150px 43px},clip]{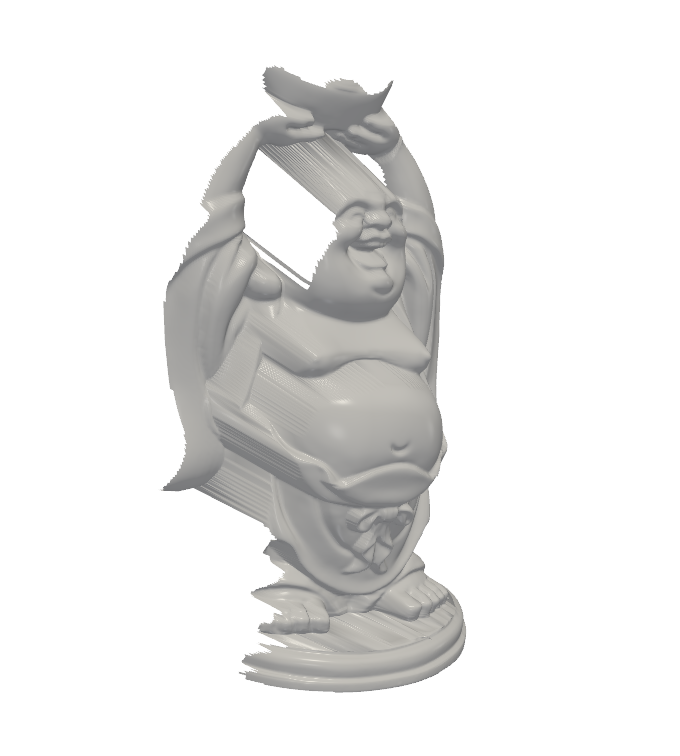}}
			&{\includegraphics[ width=\imgsize\textwidth, trim={112px 23px 112px 40px},clip]{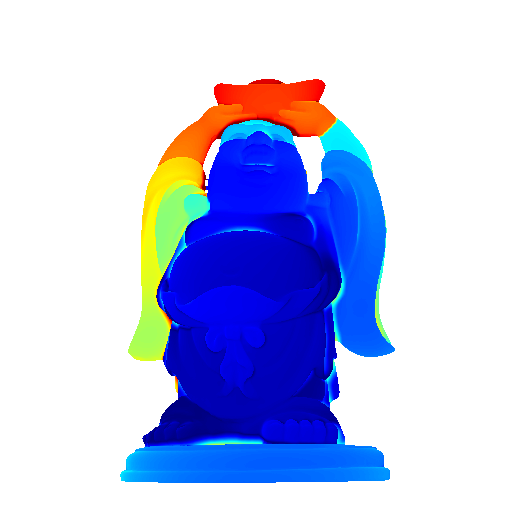}}
			&{\includegraphics[ width=\imgsize\textwidth, trim={112px 23px 112px 40px},clip]{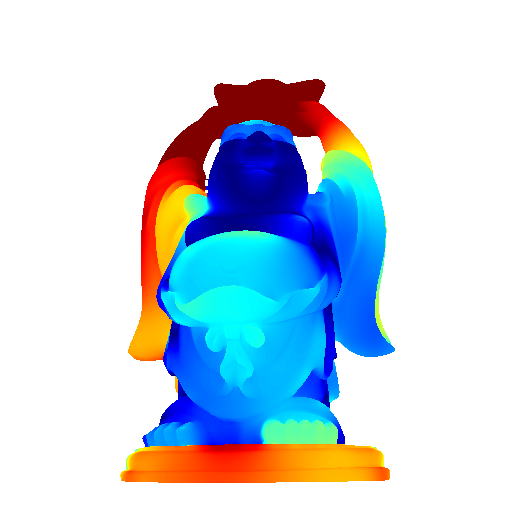}}
			&{\includegraphics[ width=\imgsize\textwidth, trim={112px 23px 112px 40px},clip]{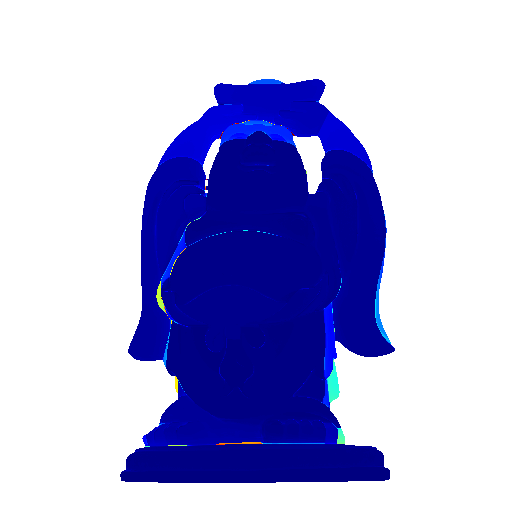}}
			\\
			\cdashline{1-7}
			GT Normal
			&\multicolumn{3}{c}{Surface normal estimation}   &\multicolumn{3}{:c}{Surface normal angular error} 
			\\
			{\includegraphics[ width=\imgsize\textwidth, trim={65px 65px 65px 65px},clip]{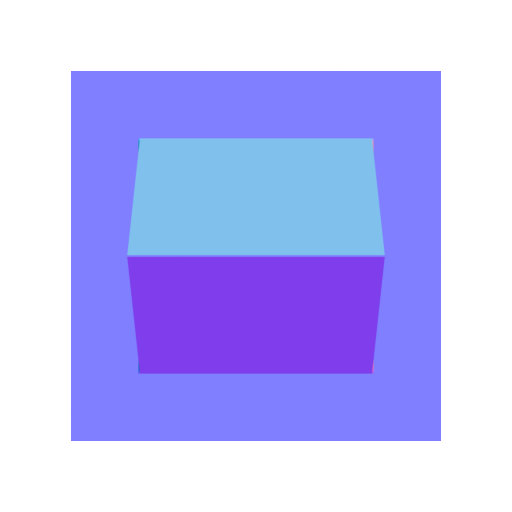}}
			&{\includegraphics[ width=\imgsize\textwidth, trim={65px 65px 65px 65px},clip]{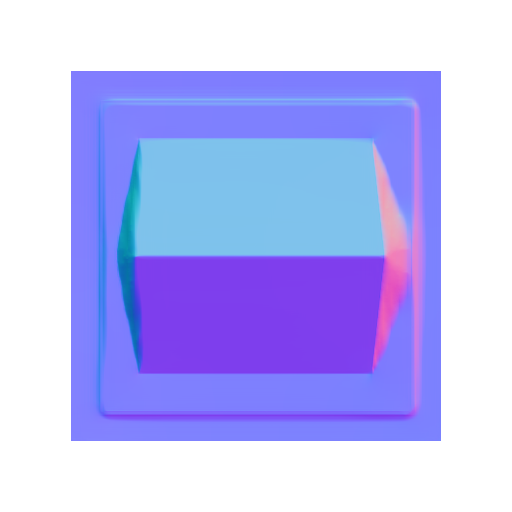}}
			&{\includegraphics[ width=\imgsize\textwidth, trim={65px 65px 65px 65px},clip]{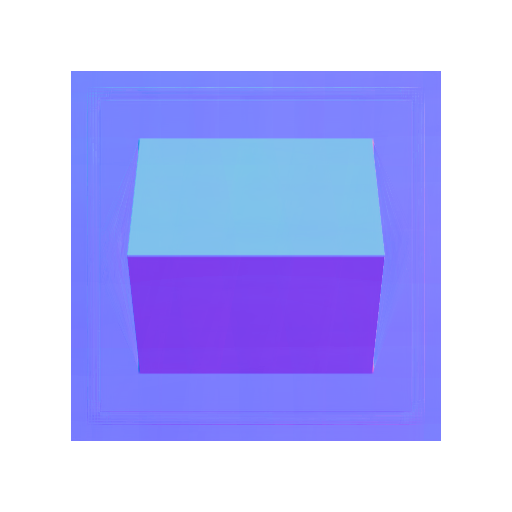}}
			&{\includegraphics[ width=\imgsize\textwidth, trim={65px 65px 65px 65px},clip]{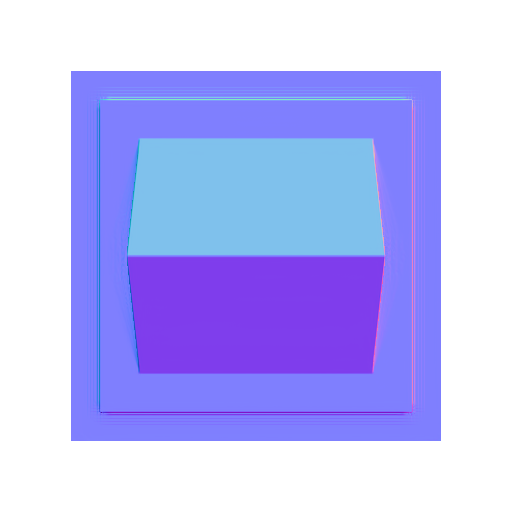}}
			&{\includegraphics[ width=\imgsize\textwidth, trim={65px 65px 65px 65px},clip]{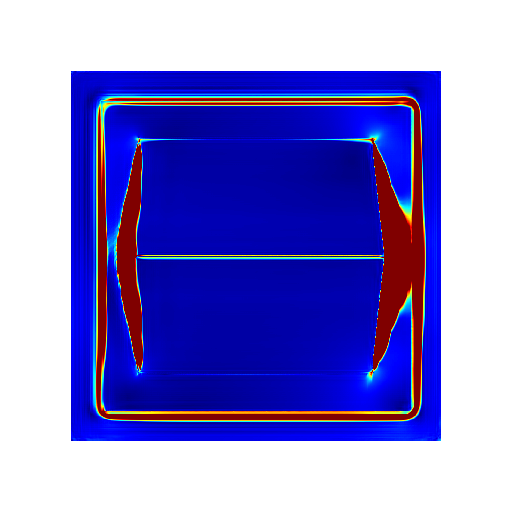}}
			&{\includegraphics[ width=\imgsize\textwidth, trim={65px 65px 65px 65px},clip]{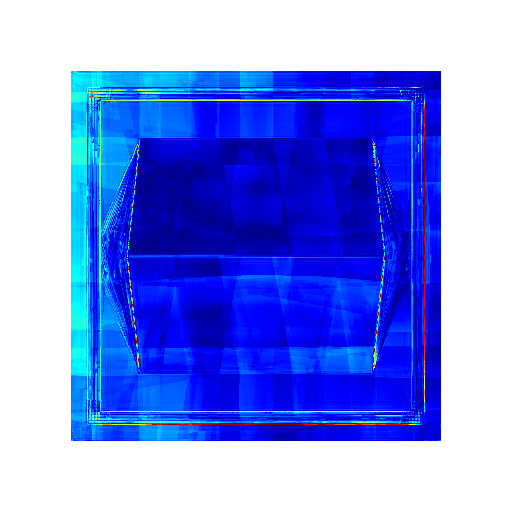}}
			&{\includegraphics[ width=\imgsize\textwidth, trim={65px 65px 65px 65px},clip]{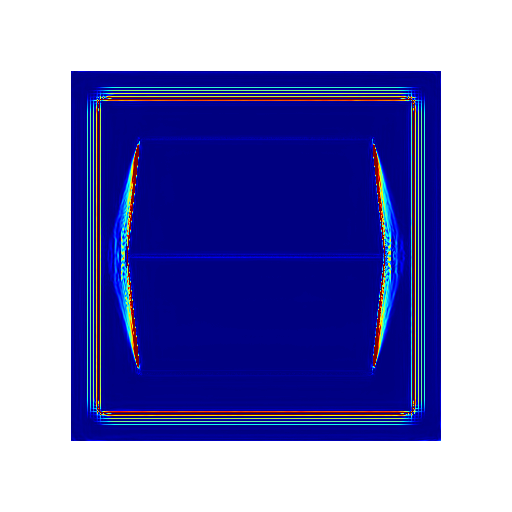}}
			\\
			{\includegraphics[ width=\imgsize\textwidth, trim={10px 0px 40px 33px},clip]{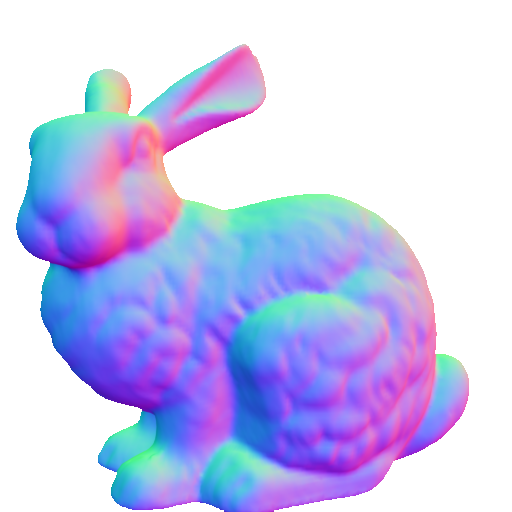}}
			&{\includegraphics[ width=\imgsize\textwidth, trim={10px 0px 40px 33px},clip]{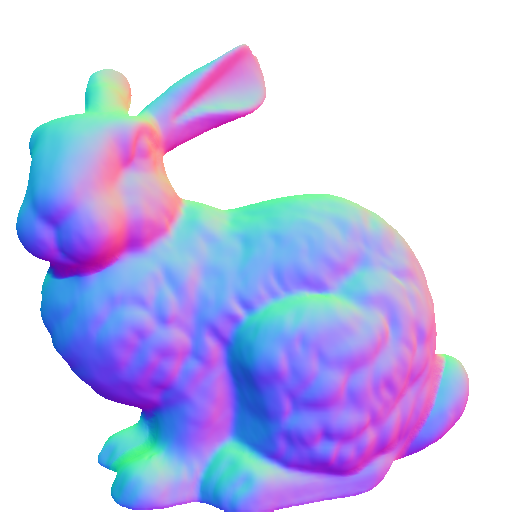}}
			&{\includegraphics[ width=\imgsize\textwidth, trim={10px 0px 40px 33px},clip]{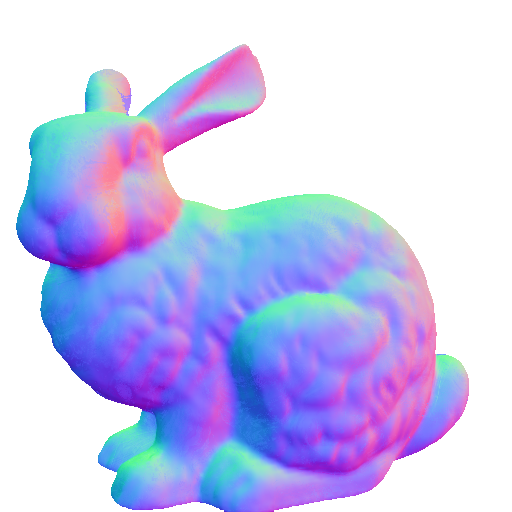}}
			&{\includegraphics[ width=\imgsize\textwidth, trim={10px 0px 40px 33px},clip]{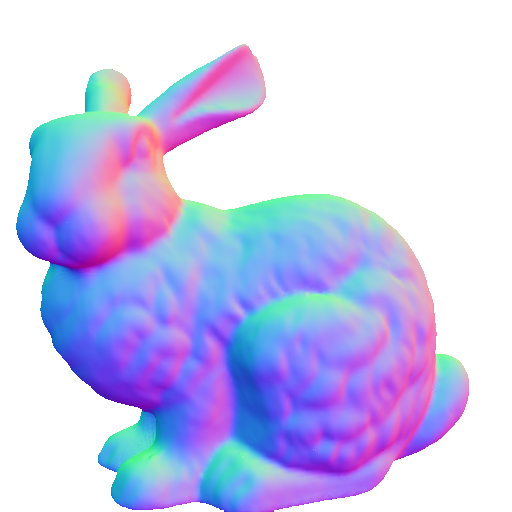}}
			&{\includegraphics[ width=\imgsize\textwidth, trim={10px 0px 40px 33px},clip]{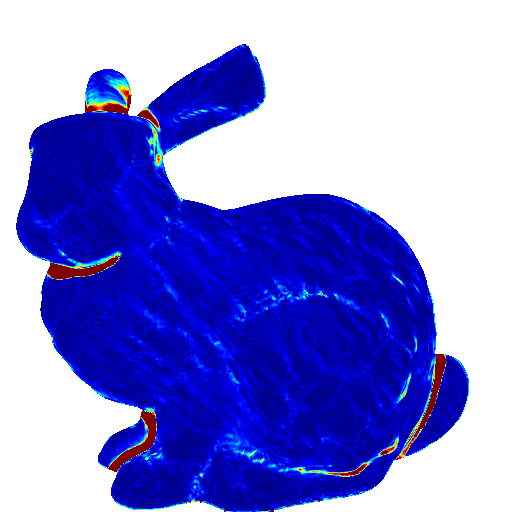}}
			&{\includegraphics[ width=\imgsize\textwidth, trim={10px 0px 40px 33px},clip]{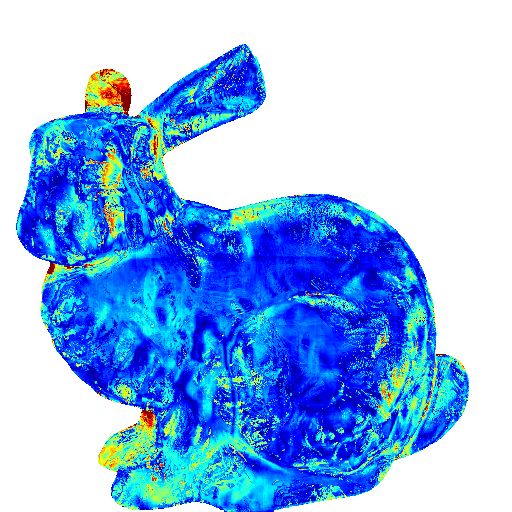}}
			&{\includegraphics[ width=\imgsize\textwidth, trim={10px 0px 40px 33px},clip]{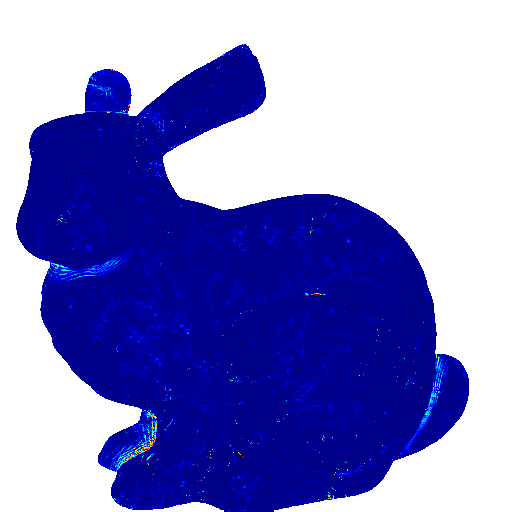}}
			\begin{minipage}{0.013\textwidth} \centering
				\vspace{-5.5em} \makebox[0.5\textwidth]{ $\,10^{\circ}$}\\ \vspace{-0em}
				\makebox[0.5\textwidth]{}\includegraphics[width=0.9\linewidth]{images/color_bar} \\ \vspace{-0.4em}
				\makebox[0.5\textwidth]{ $\,0^{\circ}$}\\
			\end{minipage}\\
			{\includegraphics[ width=\imgsize\textwidth, trim={112px 23px 112px 40px},clip]{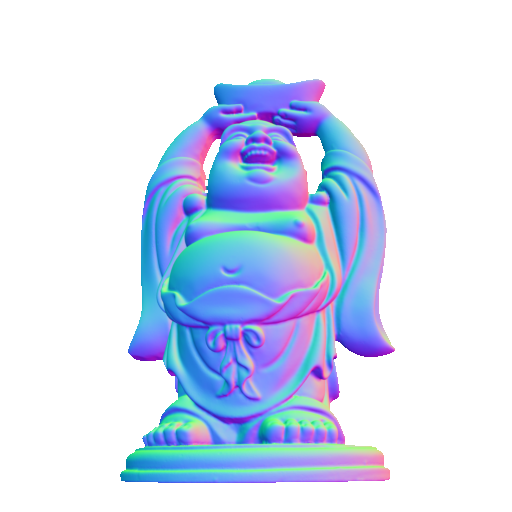}}
			&{\includegraphics[ width=\imgsize\textwidth, trim={112px 23px 112px 40px},clip]{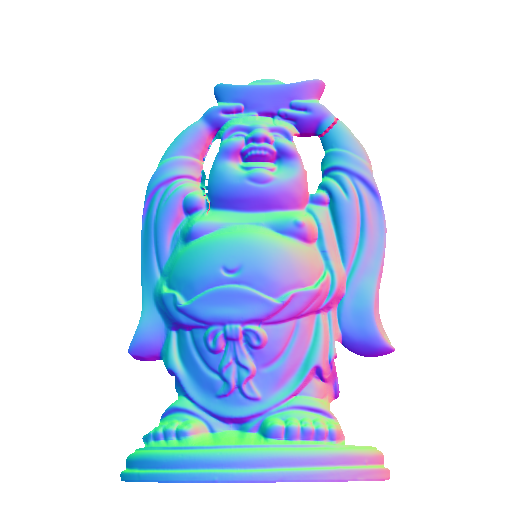}}
			&{\includegraphics[ width=\imgsize\textwidth, trim={112px 23px 112px 40px},clip]{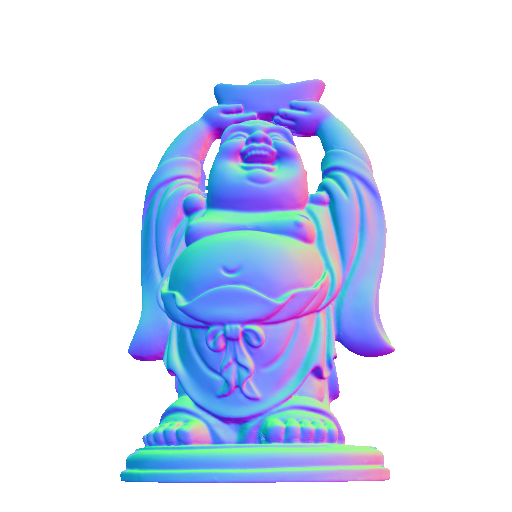}}
			&{\includegraphics[ width=\imgsize\textwidth, trim={112px 23px 112px 40px},clip]{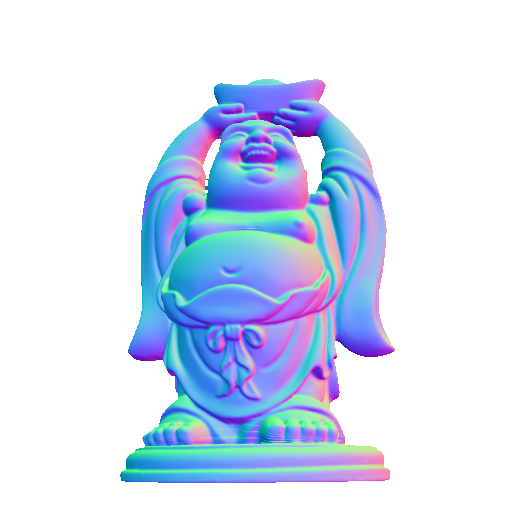}}
			&{\includegraphics[ width=\imgsize\textwidth, trim={112px 23px 112px 40px},clip]{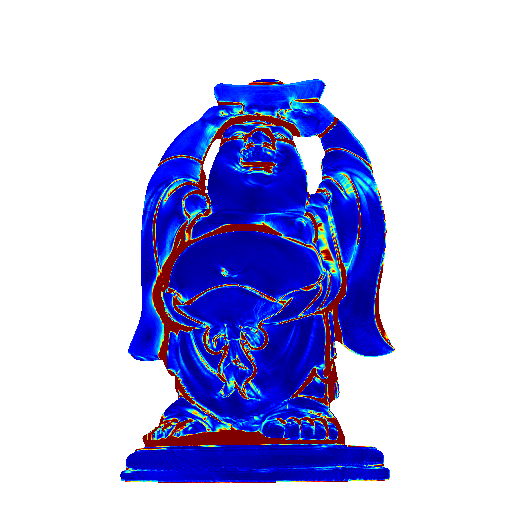}}
			&{\includegraphics[ width=\imgsize\textwidth, trim={112px 23px 112px 40px},clip]{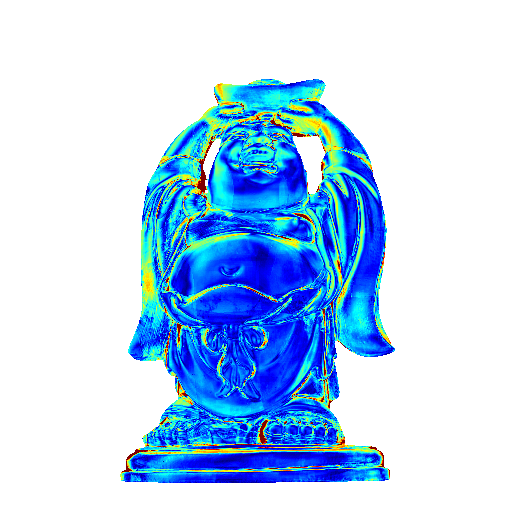}}
			&{\includegraphics[ width=\imgsize\textwidth, trim={112px 23px 112px 40px},clip]{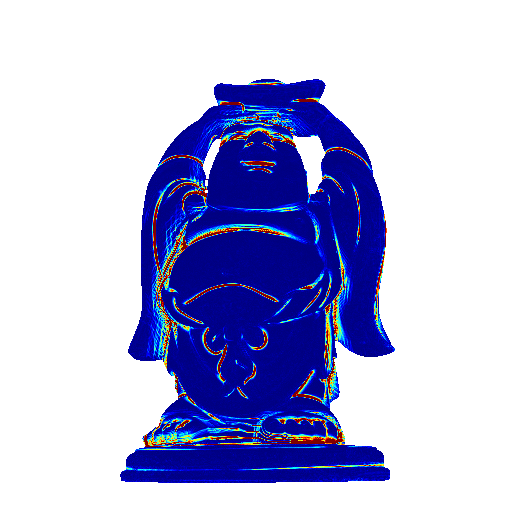}}
			\\
			&QD18~\cite{queau2018led}   &LB20~\cite{logothetis2020cnn} & Ours 			&QD18~\cite{queau2018led}   &LB20~\cite{logothetis2020cnn} & Ours\\
		\end{tabular}	
	}
	\caption{Comparison with QD18~\cite{queau2018led} and LB20~\cite{logothetis2020cnn} on synthetic data. Left side shows the GT and estimated shapes and surface normals. The right side gives the corresponding error distributions. Results and error distributions of SM20~\cite{santo2020deep} and MQ16~\cite{mecca2016single} are provided in supplementary materials.
	}
	\label{fig:synthetic_res}
	\vspace{-2em}
\end{figure*}

The comparison with the baseline methods is shown in \fref{fig:synthetic_res}. From the ground-truth (GT) shape, we can observe that the stair region in the {\sc Tent}, the ear part of the {\sc Bunny}, and the arm and foot areas of the {\sc Buddha} contain depth discontinuities. 
Our method shows the strength of depth edge preservation in such regions compared to the baseline methods, and yields shapes that are close to GTs.
From {\sc Bunny}'s error distribution, we observe that the estimation error around the ear region does not influence the shape recovery at the surrounding area with our method. 
As summarized in \Tref{table:syn_eval}, our method achieves smaller mean angular errors~(MAngE) of the surface normal and the mean absolute errors~(MAbsE) of the depth compared with existing NLPS methods. 


\begin{table}
	\centering
	\caption{Summary of depth and surface normal estimation errors.}
	\resizebox{0.95\textwidth}{!}{
	\begin{threeparttable}
			\begin{tabular}{ccccccccccc}
				\toprule
				\multirow{2}{*}{Object} & \multicolumn{2}{c}{QD18~\cite{queau2018led}} & \multicolumn{2}{c}{LB20~\cite{logothetis2020cnn}} & \multicolumn{2}{c}{SM20~\cite{santo2020deep}} & \multicolumn{2}{c}{MQ16~\cite{mecca2016single}} & \multicolumn{2}{c}{Ours} \\
				\cmidrule(l){2-3} \cmidrule(l){4-5} \cmidrule(l){6-7} \cmidrule(l){8-9} \cmidrule(l){10-11}
				& MAngE\tnote{1} & MAbsE\tnote{2} &  MAngE & MAbsE& MAngE & MAbsE & MAngE & MAbsE & MAngE & MAbsE \\
				\midrule
				{\sc Tent} & 5.41 & 55.2 & 2.29 & 93.5  & 9.99 & 371.1 & 10.99 & 111.3 & 1.72 & 2.68 \\
				{\sc Bunny} & 1.66 & 24.1 & 3.94 & 30.8  & 7.43 & 57.9 & 8.54 & 60.0 & 0.33 & 2.73 \\
				{\sc Buddha} & 7.84 & 70.9 & 4.44 & 129.1 & 12.19 & 149.2 & 10.44 & 97.2 & 2.11 & 9.00 \\
				Average & 4.97 & 50.1 & 3.57 & 84.5 & 9.87 & 192.7 & 9.98 & 89.5 & \textbf{1.39} & \textbf{4.80} \\
				\bottomrule
			\end{tabular}
		\begin{tablenotes}[para]
			\item[1] MAngE unit: Degree
			\item[2] MAbsE unit: mm
		\end{tablenotes}
	\end{threeparttable}
	}
	\label{table:syn_eval}
\end{table}

\begin{figure}
	\resizebox{1\textwidth}{!}{
	\begin{tabular}{c@{}@{}c@{}@{}c@{}@{}c@{}:@{}c@{}@{}c@{}@{}c@{}@{}c@{}@{}c@{}@{}c}
		42.5 & 9.81 & 46.2 & 44.2 &2.68 & 2.73 & 9.0 & 29.6 \\
		{\includegraphics[ height=\imgsize\textwidth, trim={375px 54px 341px 78px},clip]{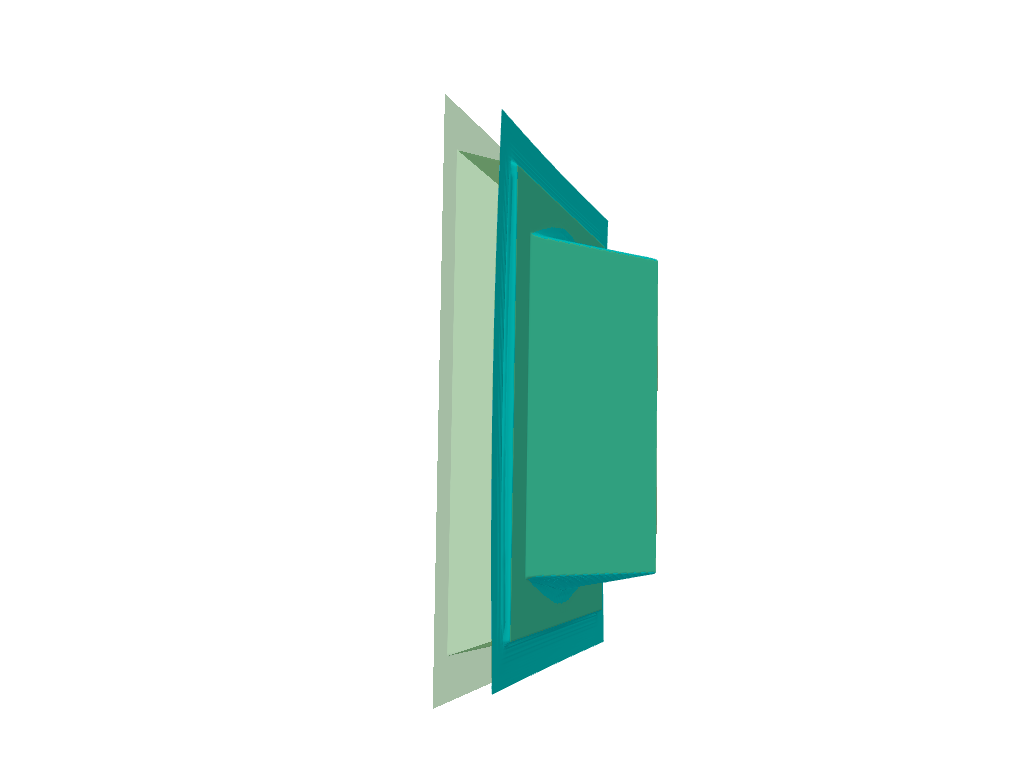}}
		&{\includegraphics[ height=\imgsize\textwidth, trim={230px 88px 325px 129px},clip]{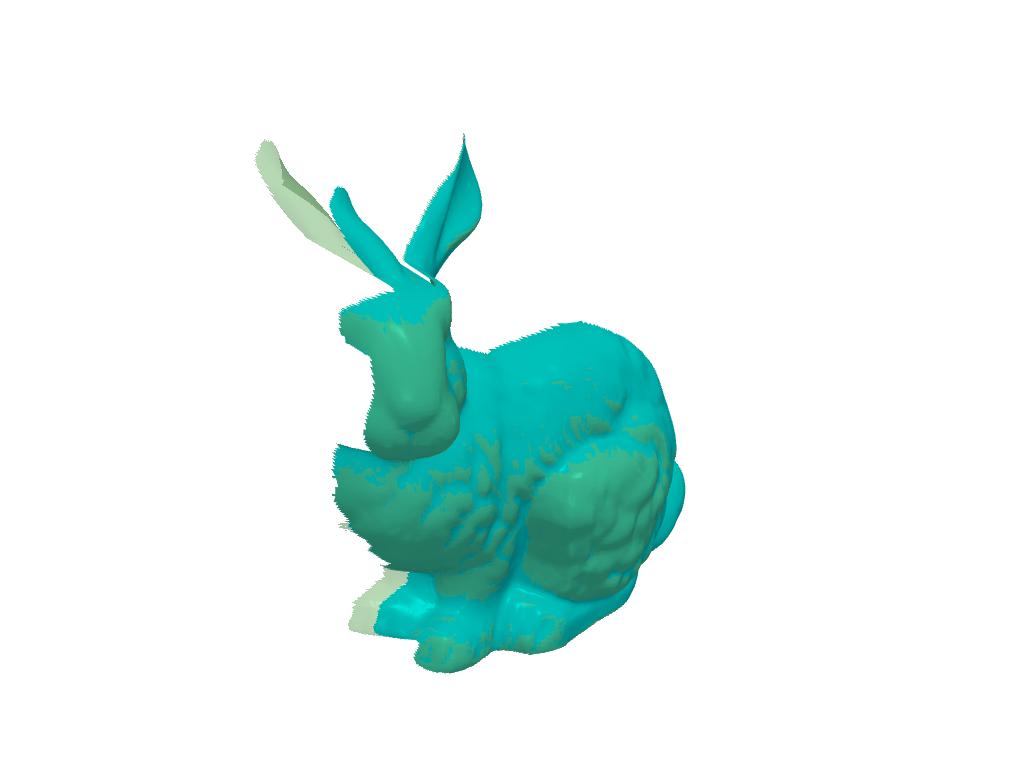}}
		&{\includegraphics[ height=\imgsize\textwidth, trim={275px 62px 367px 54px},clip]{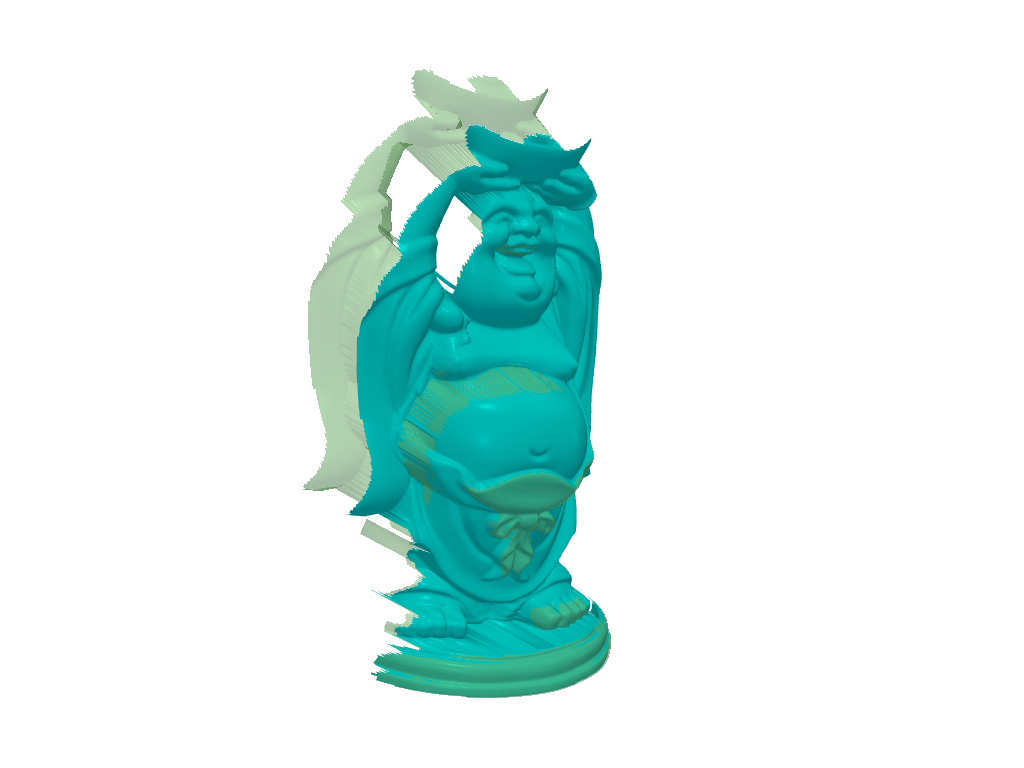}}
		&{\includegraphics[  height=\imgsize\textwidth, trim={381px 73px 268px 87px},clip]{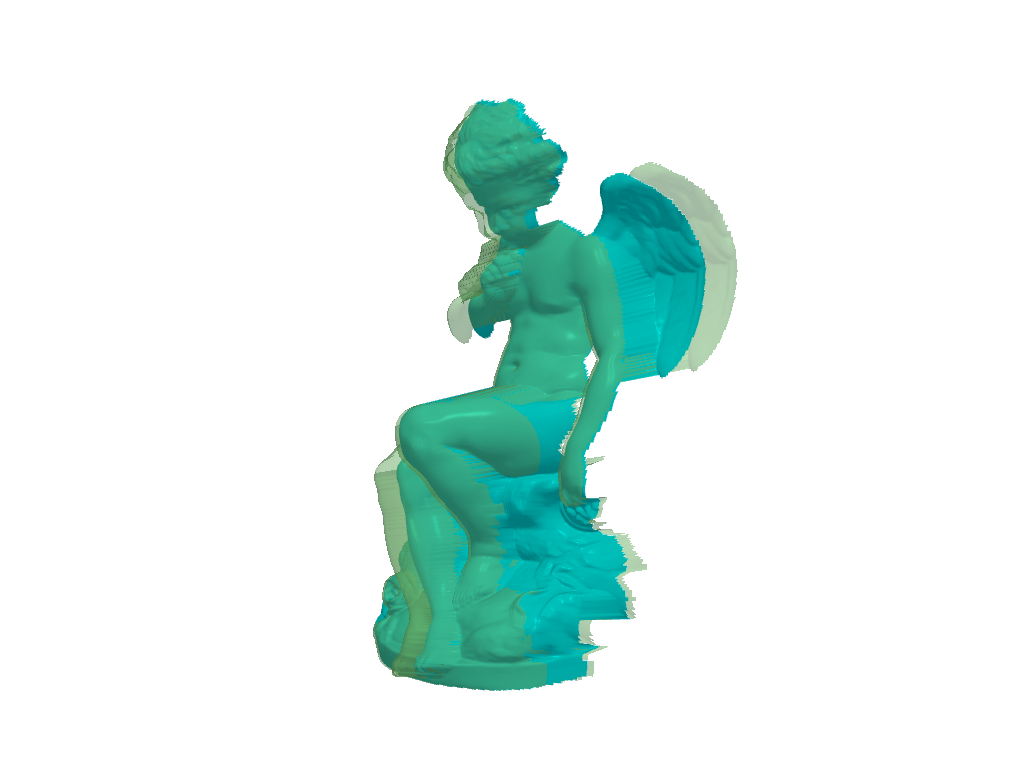}} 
		&{\includegraphics[ height=\imgsize\textwidth, trim={375px 54px 341px 78px},clip]{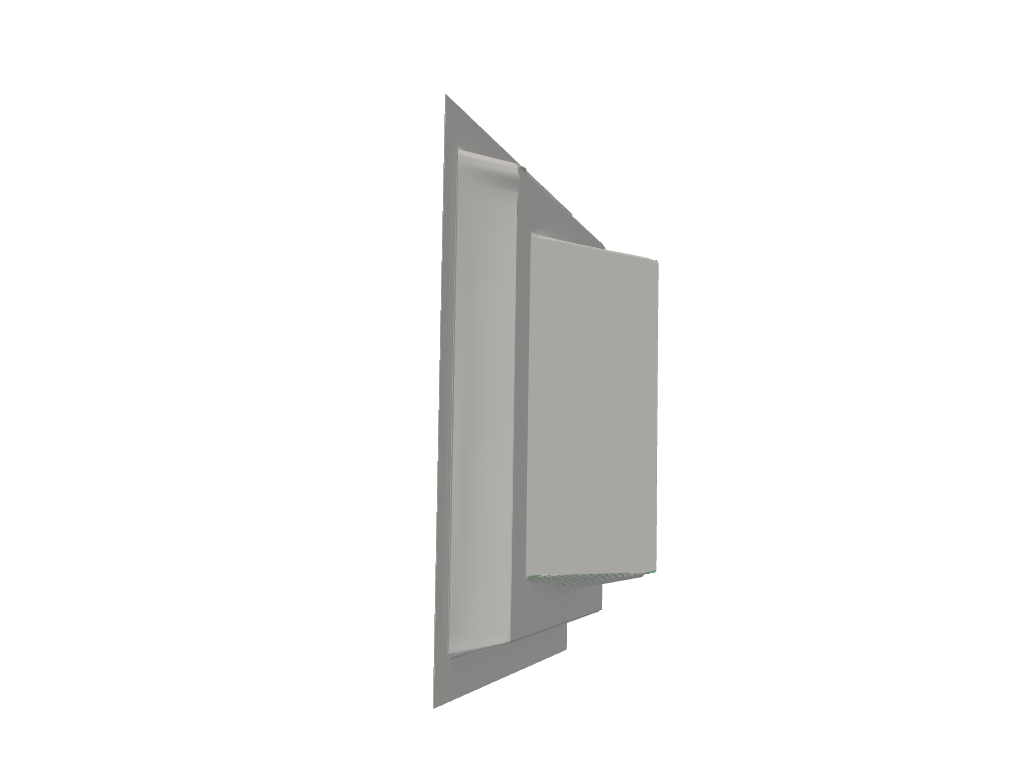}}
		&{\includegraphics[ height=\imgsize\textwidth, trim={230px 88px 325px 129px},clip]{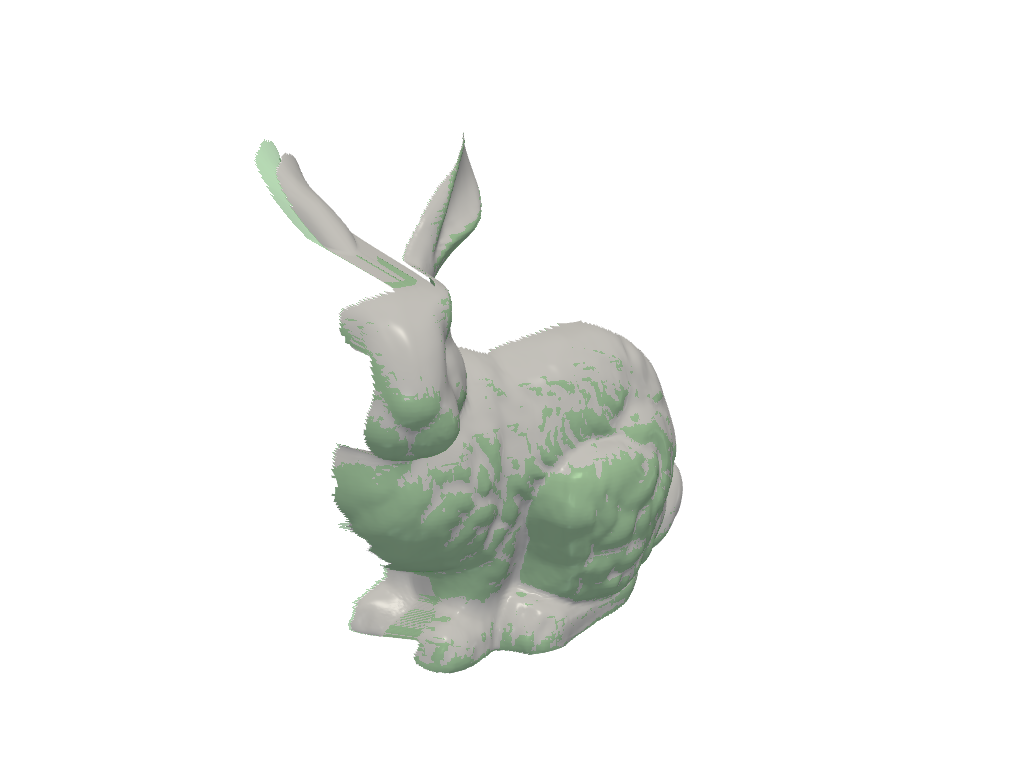}}
		&{\includegraphics[ height=\imgsize\textwidth, trim={275px 62px 367px 54px},clip]{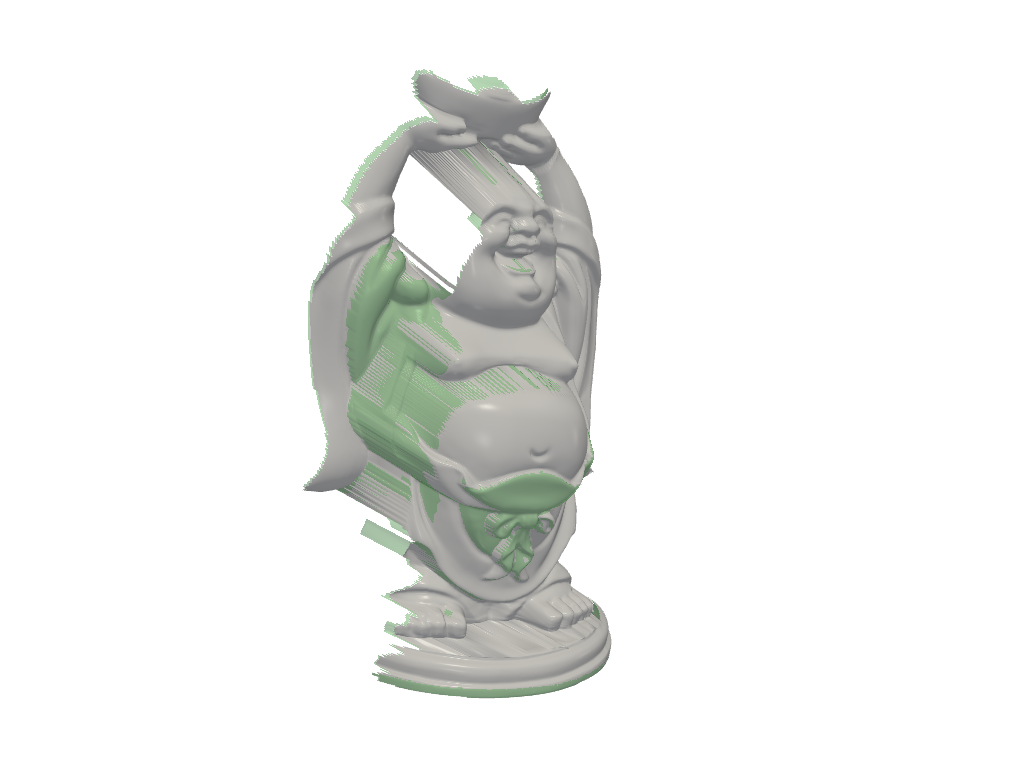}}
		&{\includegraphics[ height=\imgsize\textwidth, trim={381px 73px 268px 87px},clip]{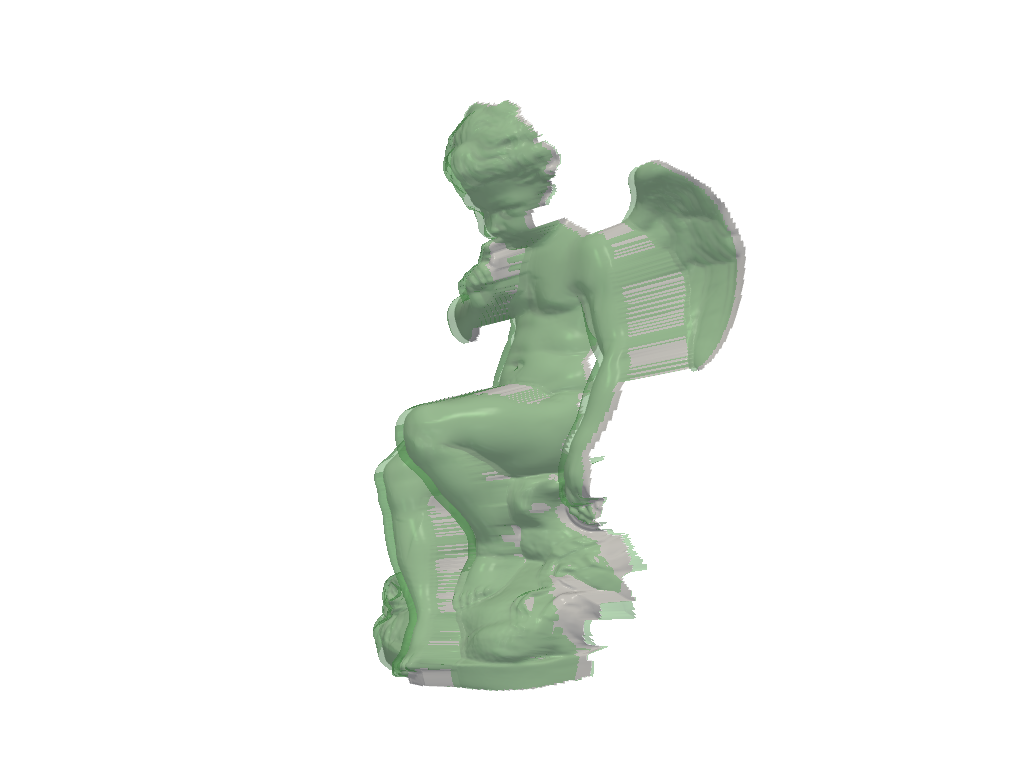}}\\
	\end{tabular}
}
\\
\centering
{\includegraphics[ width=0.6\textwidth]{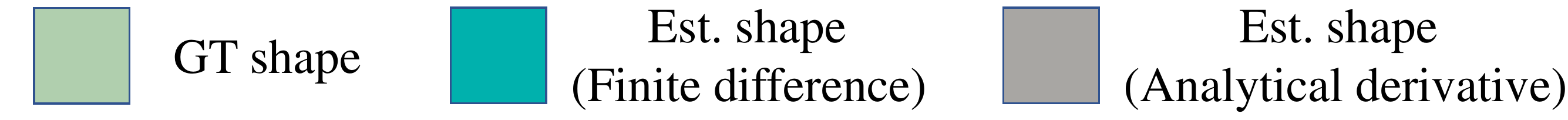}}

	\vspace{-0.2em}
	\caption{Reconstructed shapes by our method with finite difference~(left) and analytical derivative~(right). Sharp edges are well-preserved using analytical derivatives. The depth MAbsE in mm is shown on the top of each shape estimate. 
	}
	\label{fig:finite_diff_ablation}
\end{figure}

\paragraph{Effectiveness of analytical derivatives}
To assess the effectiveness of the analytical differentiation over finite differentiation, we compare our method with the version where the analytical derivatives are replaced by finite differences calculated from depth estimates in each forward pass. 
The comparison is summarized in \fref{fig:finite_diff_ablation}.
Compared to our method with the analytical derivatives, the method with finite differences shows greater errors. In particular, our method based on analytical derivatives shows faithful recovery around the depth discontinuities.
\begin{figure}
	\vspace{-1em}
	\centering
	\begin{tabular}{@{}c@{}@{}c@{}@{}c@{}}
	{\includegraphics[ height=0.21\textwidth]{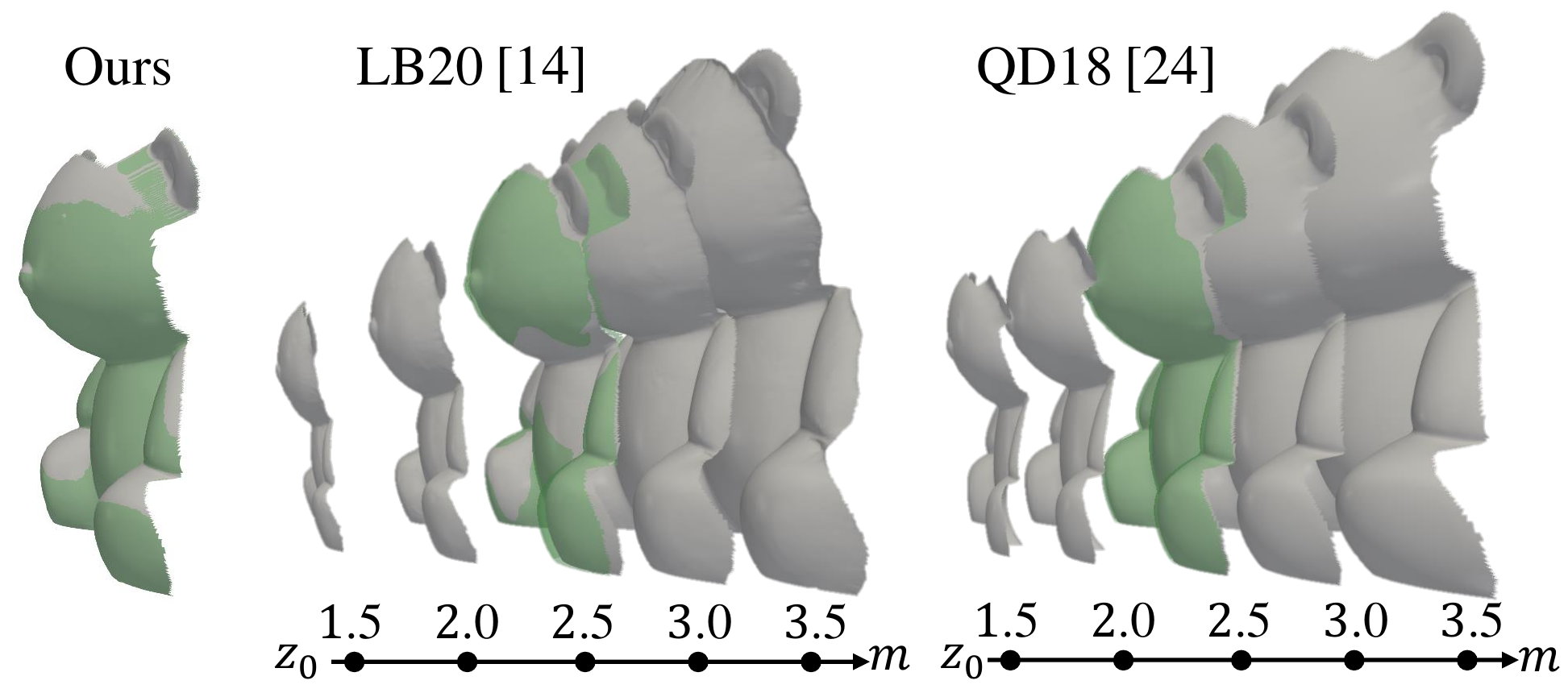}} 
	& {\includegraphics[ height=0.21\textwidth]{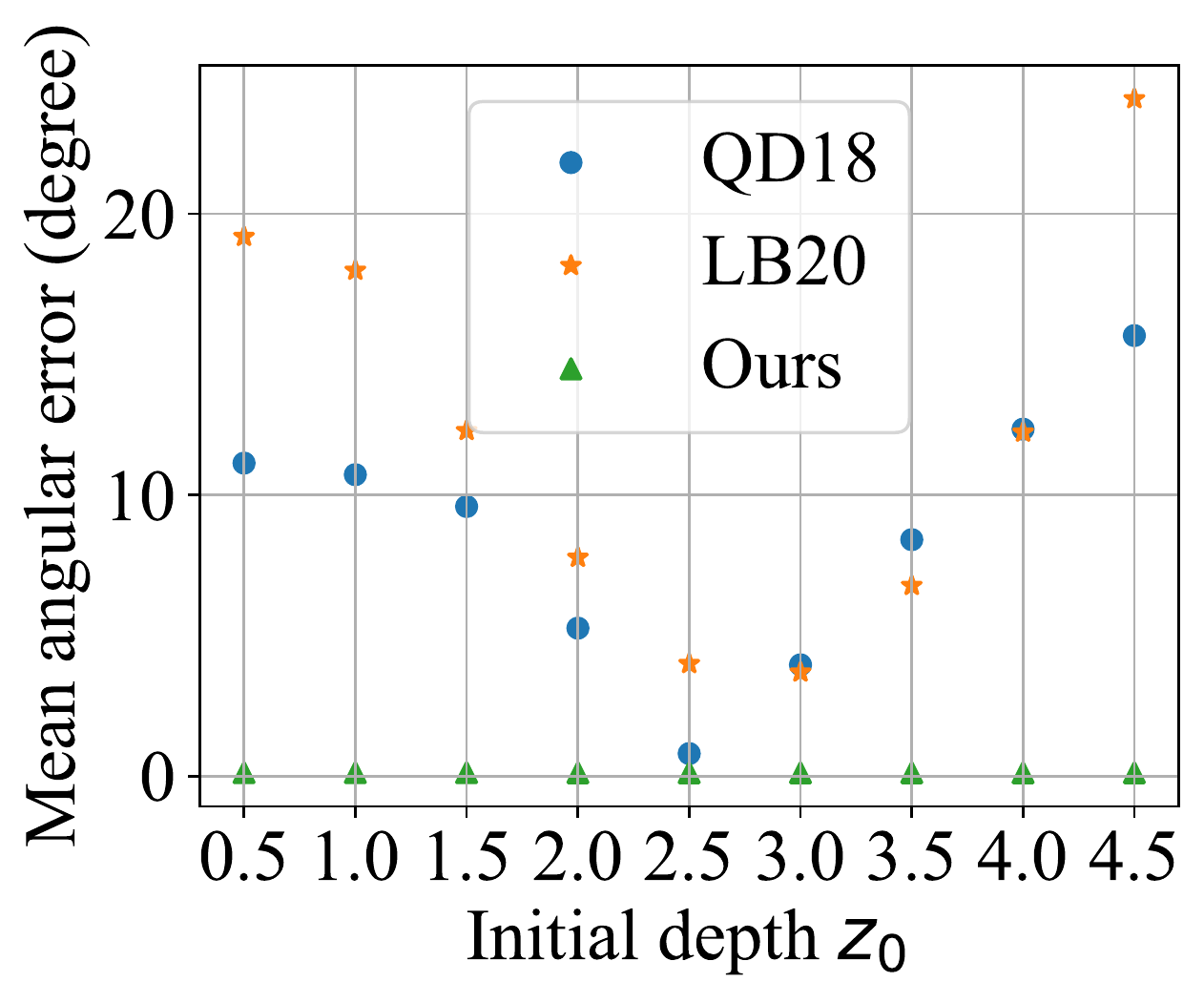}} 
	 &{\includegraphics[ height=0.21\textwidth]{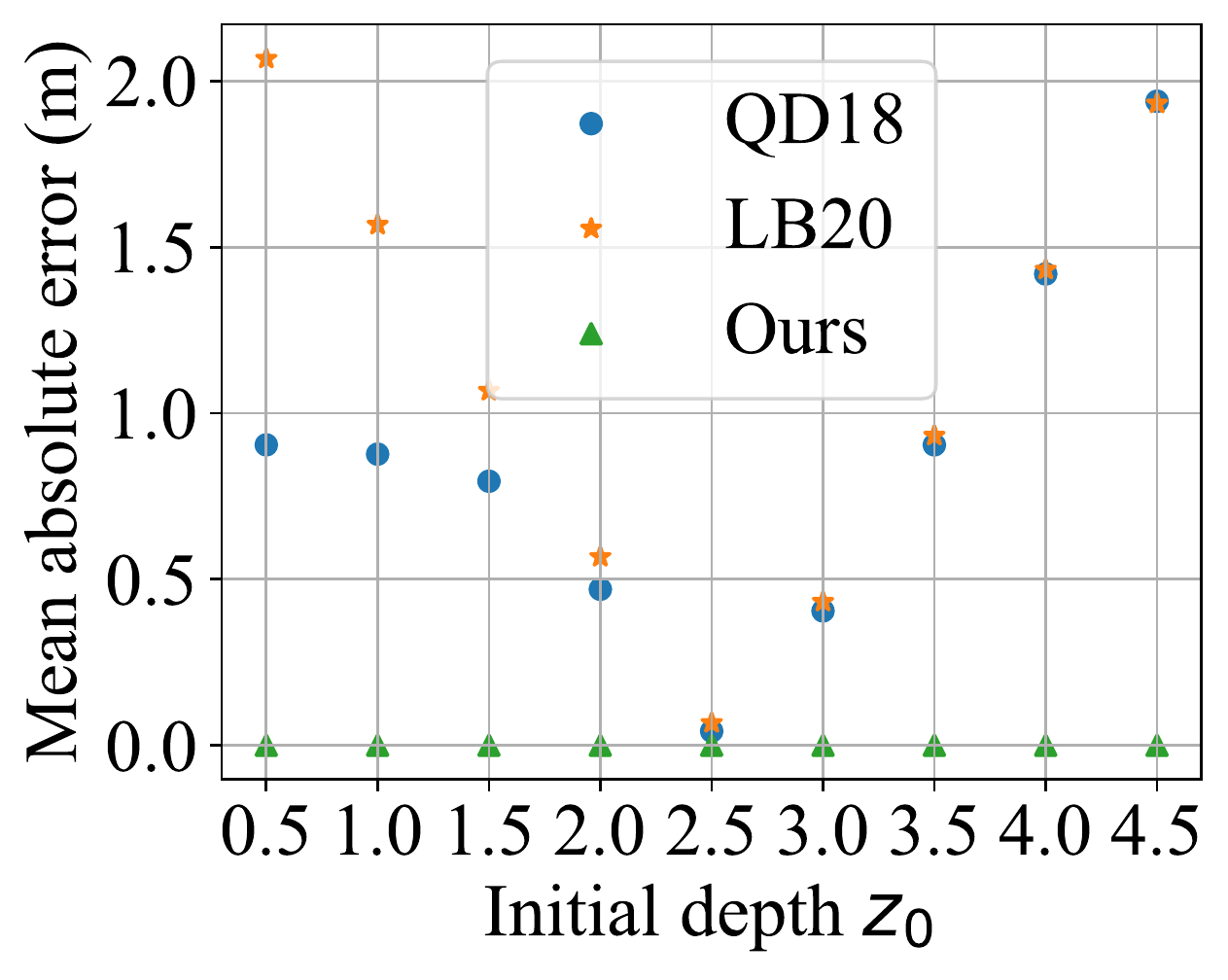}}
\end{tabular}

	\caption{Evaluation of the sensitivity to varying initial depths. (\textbf{Left}) Recovered shapes of our and baseline methods under varying initial depths, where the GT shape is shown in light green. The result of our method is unchanged regardless of the initial depths in this range. (\textbf{Right}) Quantitative evaluations of the surface normal and depth under different initial depth guesses $z_0$.}
	\label{fig:initial_eval}
	\vspace{-1.5em}
\end{figure}

\paragraph{Robustness against depth initialization}
In our method, the initial shape is determined by the initialization of network parameters. We use the initialization strategy of the {\sc Siren}~\cite{sitzmann2020implicit} network, with which the initial depth follows a normal distribution centered at an offset.
The initial depth offset is controlled by the initial bias value of the output layer/neurons.
To evaluate the robustness of our method against the depth initialization, we vary the initial depth offset $\{z_0\}$ and evaluate the shapes recovered by the baseline and our methods. To eliminate the influence of discontinuities to the baseline methods, we choose a smooth shape in this experiment. 

Under the varying initial depths in a relatively large range, our method always converges to the same shape estimation, which consistently overlaps with the GT shape as shown in \fref{fig:initial_eval}.
From the estimation error of surface normal and depth shown in the plots of \fref{fig:initial_eval}, our method is not sensitive to the initial shape as our method jointly optimizes the surface shape and albedo by only updating the neural surface parameters.
Compared to the ground truth, the recovered shapes from LB20~\cite{logothetis2020cnn} and QD18~\cite{queau2018led} become either flat~($z_0 = 1.5\,$m) or inflating~($z_0 = 3.5\,$m) depending on the initial depth settings. 
When the depth is initialized as the mean distance between the target object and the camera~($z_0 = 2.5\,$m in the experiment), the baseline methods achieve the high accuracy in the surface normal and depth estimation.

\renewcommand\imgsize{0.1}
\begin{figure*}
		\resizebox{0.41\textwidth}{!}{
					\small
		\begin{tabular}{@{}cc@{}@{}c@{}}
			\toprule
			\makecell[c]{(A) Lambertian Diffuse  \\ $\sigma_d = 0$, $\sigma_s = 0$}
			& { \begin{minipage}{\imgsize\textwidth}
					\includegraphics[ width=\textwidth]{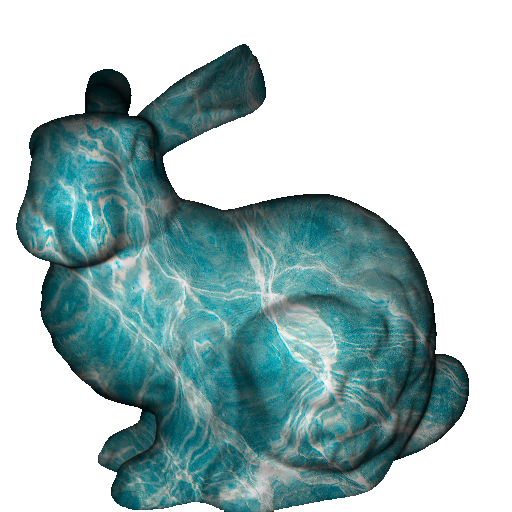}
				\end{minipage}} 
			& { \begin{minipage}{\imgsize\textwidth}
					\includegraphics[ width=\textwidth]{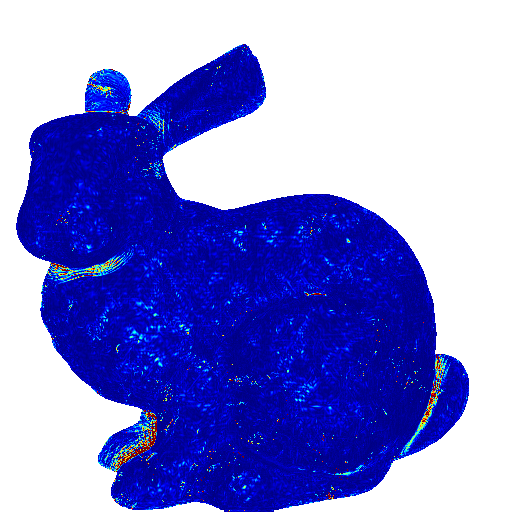}
			\end{minipage}} 
			\\
			\makecell[c]{(B) Oren-Nayar Diffuse \\ $\sigma_d = 1$, $\sigma_s = 0$}
			& { \begin{minipage}{\imgsize\textwidth}
					\includegraphics[ width=\textwidth]{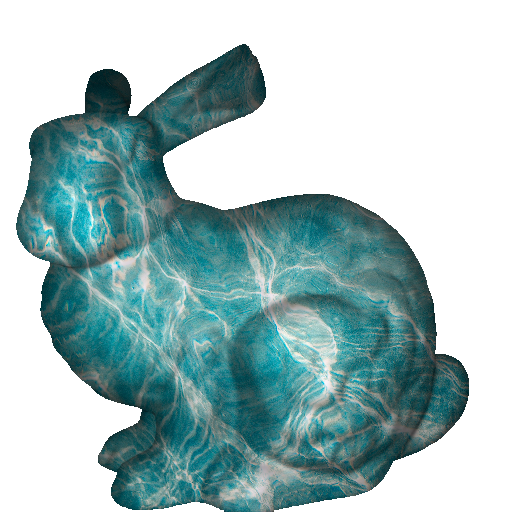}
			\end{minipage}} 
			& { \begin{minipage}{\imgsize\textwidth}
					\includegraphics[ width=\textwidth]{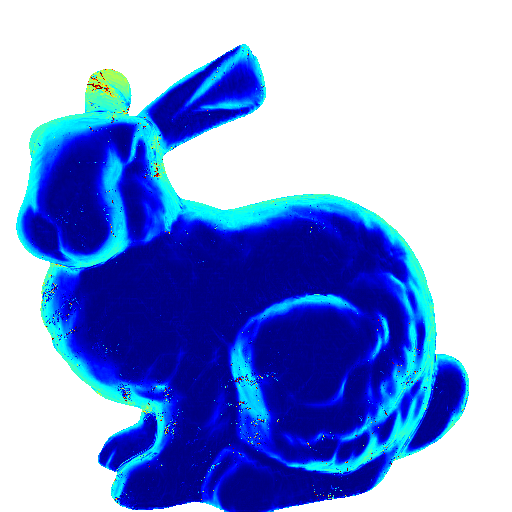}
			\end{minipage}} 			
			\\
			\makecell[c]{(C) Specular Reflectance \\ $\sigma_d = 0$ , $\sigma_s = 0.4$}
			& { \begin{minipage}{\imgsize\textwidth}
					\includegraphics[ width=\textwidth]{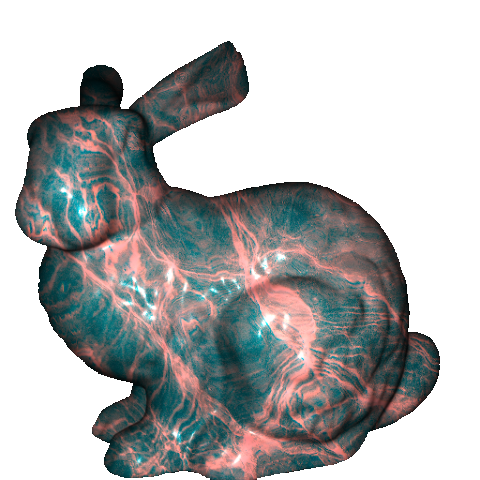}
			\end{minipage}} 
			& { \begin{minipage}{\imgsize\textwidth}
					\includegraphics[ width=\textwidth]{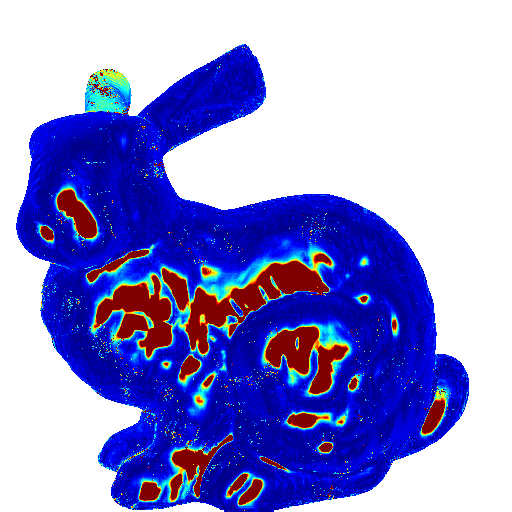}
			\end{minipage}} 
			\\
			\bottomrule
		\end{tabular}
		}
	\hspace{-1em}
	\begin{minipage}{0.01\textwidth} \centering
		\vspace{1em} \makebox[0.6\textwidth]{\scriptsize $\,\;\, 5^{\circ}$}\\ \vspace{-0em}
		\makebox[0.4\textwidth]{}\includegraphics[width=0.4\linewidth]{images/color_bar} \\ \vspace{-0.4em}
		\makebox[0.5\textwidth]{\scriptsize $\,\,\, 0^{\circ}$}\\
	\end{minipage}
	\hfill
		\resizebox{0.58\textwidth}{!}{
			\begin{tabular}{cc}
				{\includegraphics[ height=\textwidth]{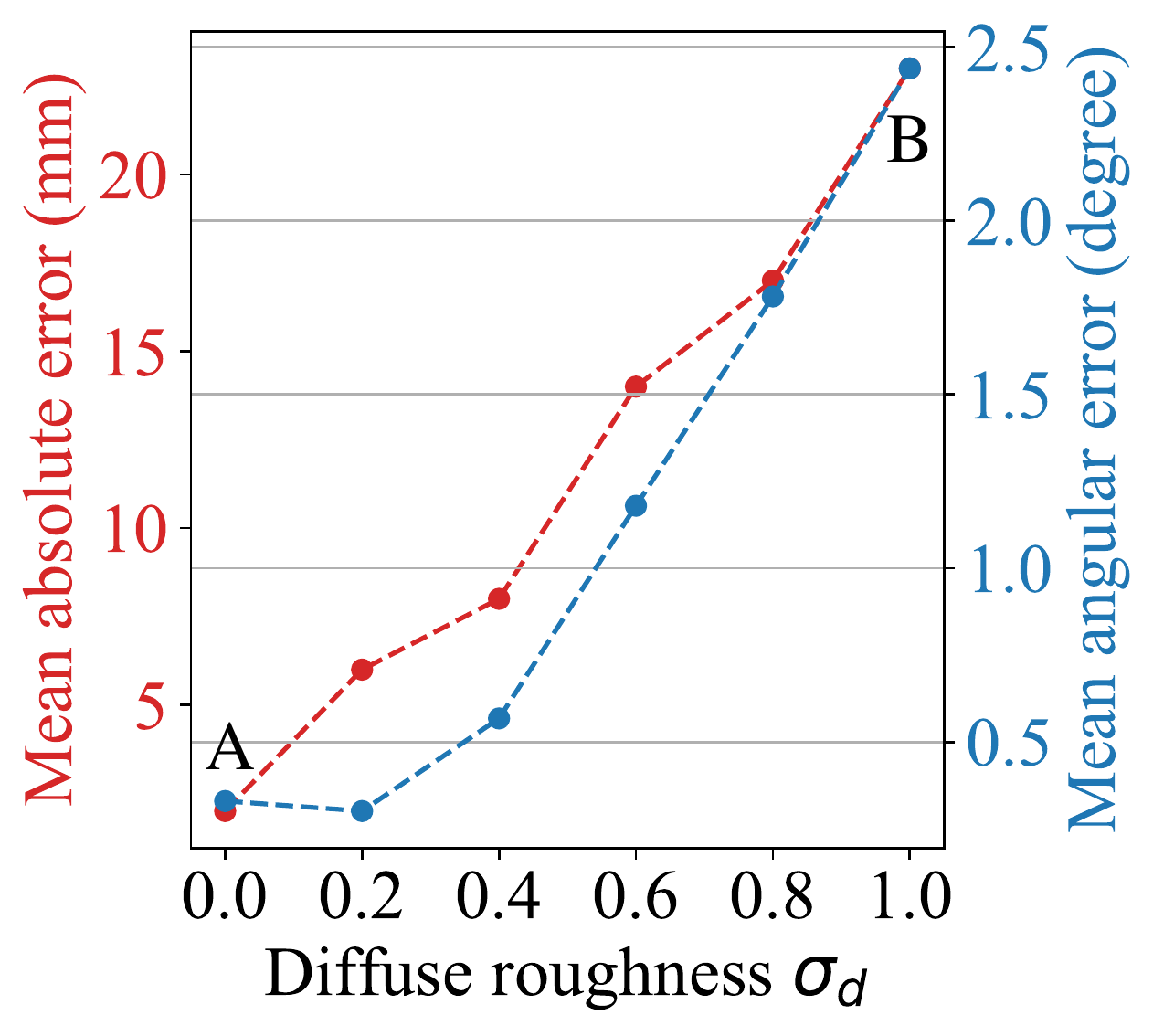}}
				&{\includegraphics[ height=\textwidth]{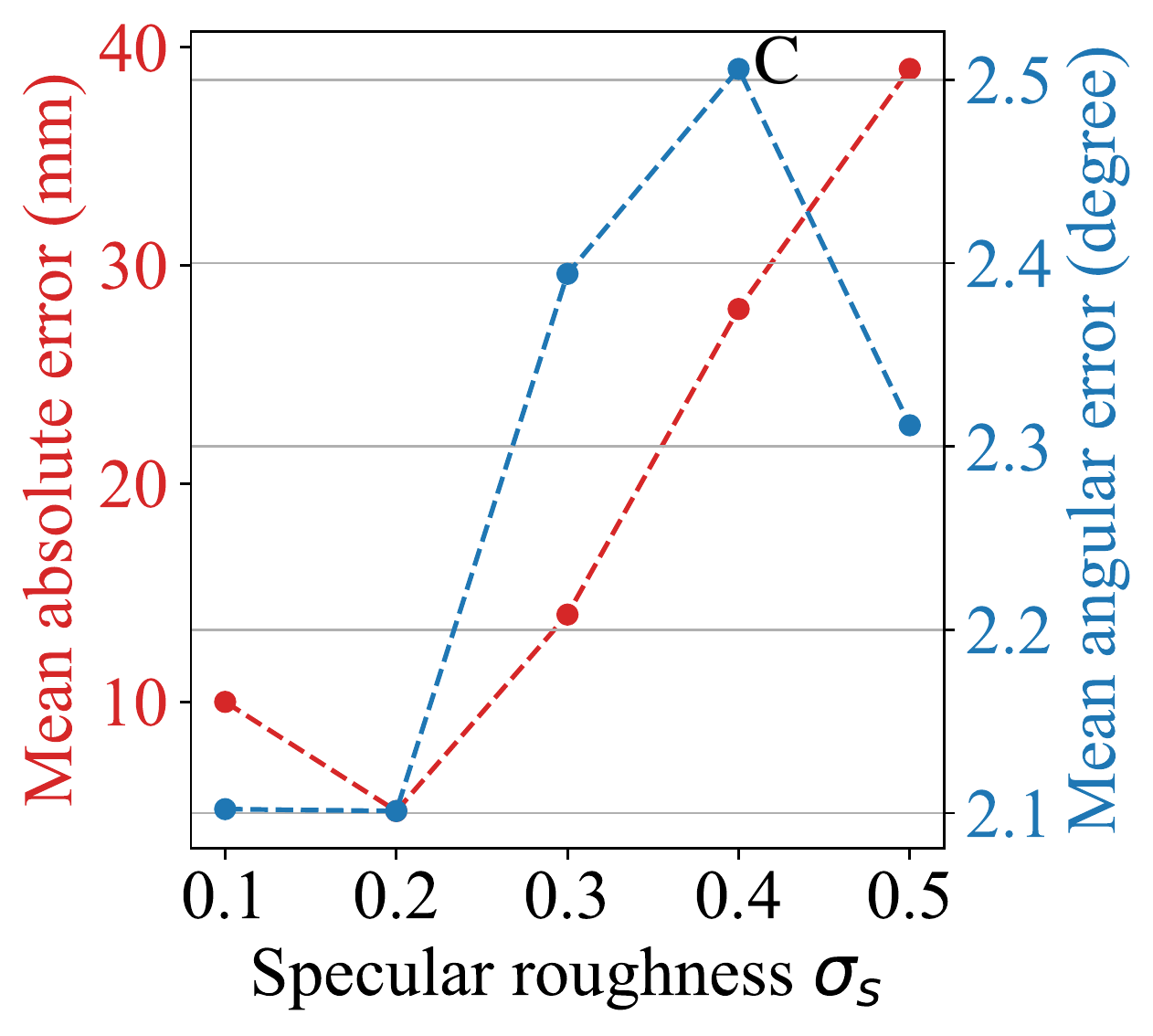}}
			\end{tabular}
			}

	\caption{Evaluation using non-Lambertian surfaces. (\textbf{Left}) Image observations and the angular error distributions under the reflectance of Lambertian diffuse, Oren–Nayar~\cite{oren1994generalization} diffuse, and the specular reflectance using Disney principled BSDF~\cite{burley2012physically}. (\textbf{Right}) Depth and surface normal estimation errors of our method under varying values of diffuse roughness $\sigma_d$ and specular roughness $\sigma_s$.}
	\label{fig:non_lambert_eval}
\end{figure*}

\paragraph{Influence of non-Lambertian reflectances}
Our NLPS method assumes Lambertian diffuse surfaces. 
To evaluate the influence of non-Lambertian reflectances, we test our method using a more general diffuse reflectance rendered by Oren–Nayar model~\cite{oren1994generalization}, and also the specular reflectance modeled by Disney principled BSDFs~\cite{burley2012physically}. We render the image observations with varying diffuse roughness $\sigma_d$ and specular roughness $\sigma_s$ values of the two reflectance models, and evaluate the influence of non-Lambertian reflectances on our method.

\Fref{fig:non_lambert_eval} (left) shows the image observations under Lambertian diffuse reflectance, Oren–Nayar reflectance, and Disney principled specular reflectance based on different settings of $\sigma_d$ and $\sigma_s$.
From the error distribution of the surface normal estimates, Oren–Nayar reflectance brings errors to surface normals with large elevation angles, and specular reflectance influences the area where specular highlights are observed.
The estimation errors of our method under varying roughnesses are shown on the right side. As expected, the surface normal and depth estimation accuracies decrease with the reflectance deviating from the Lambertian reflectance. However, the influences of the non-Lambertian reflectance are rather limited in this experiment. The surface normal estimation error increases from $0.33\degree$ to less than $2.4\degree$. The depth error increase from $2.73$~mm to $40$~mm for the {\sc Bunny} object with the size of $2$~m.

The accuracy of our NLPS method for non-Lambertian scenes has also been tested using real dataset LUCES~\cite{mecca2021luces} containing diverse materials. The results presented in the supplementary material show that our method yields reasonable shape recoveries for a set of non-Lambertian surfaces.

\subsection{Experiments on Real-world Dataset}
Besides the synthetic evaluation, we also test our method using real captured data to assess our method on real scenarios. In this section, we first introduce our hardware setup for near-light image data capture, then we compare the shape recovery results with the baseline methods.

\begin{figure*}[t]
	\centering
	\includegraphics[width=0.75\textwidth]{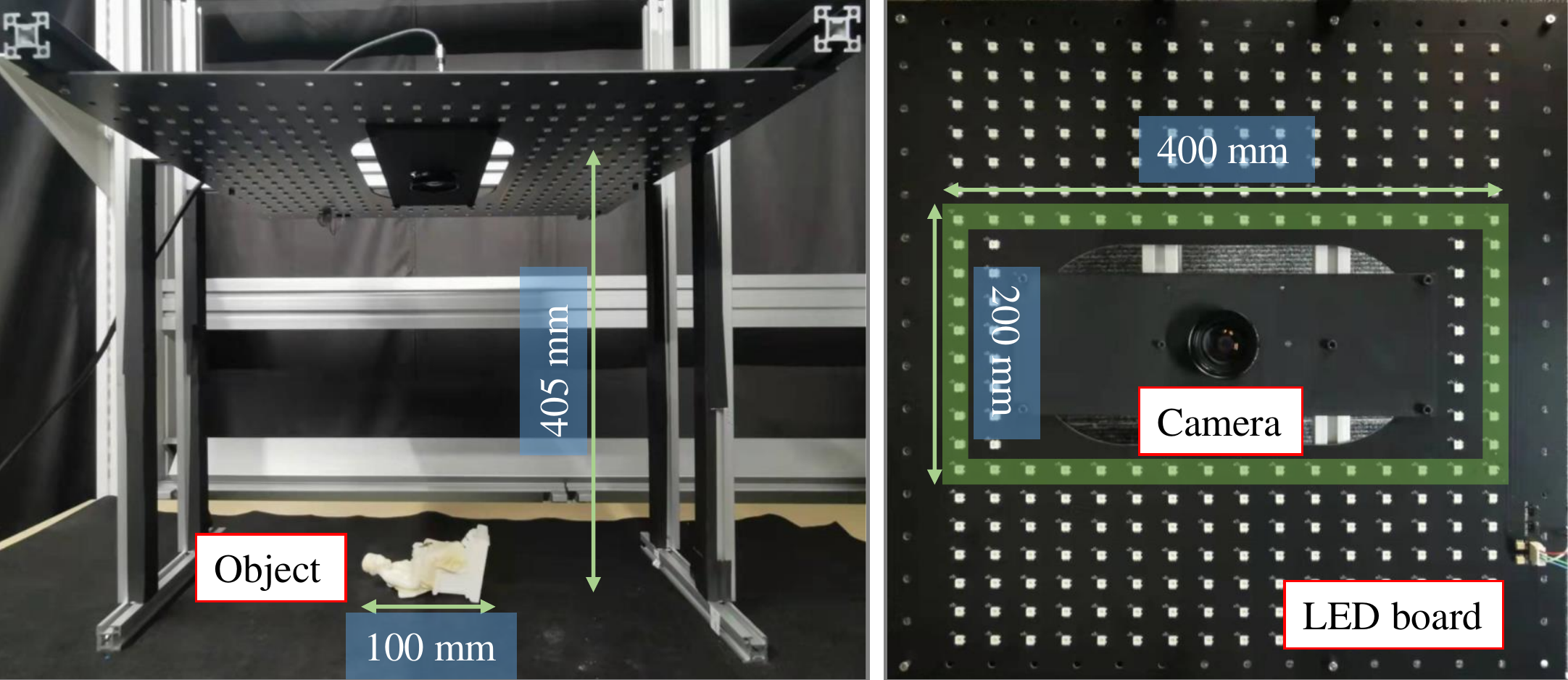}
	\caption{Hardware setup for near-light photometric stereo. The object size and the LED to object distance are designed to maintain a near light illumination. Green area indicates the LED droids used for our data capturing.}
	\label{fig:hardware_setup}
	\vspace{-1em}
\end{figure*}

\paragraph{Hardware setup} As shown in \fref{fig:hardware_setup}, we build a device for capturing near-light image observations, in which the LED positions and object size are designed to ensure the near-light setting. We calibrate the LED positions and the intrinsics of our camera~(FLIR BFS-U3-123S6C-C) with a $16$~mm lens. Since all the LEDs are of the same product\footnote{Normand SC6812RGBW LED, \url{http://www.normandled.com/Product/view/id/883.html}. Retrieved March 7th, 2022.}, we assume they share the same radiant parameters~(\hbox{$\Phi_0 = 1, \mu = 0.574$}).
\renewcommand\imgsize{0.2}

\begin{figure}
	\vspace{-1em}
	\scriptsize
	\makebox[0.12\textwidth]{\quad Near-light images}
	\makebox[0.33\textwidth]{\quad Surface normal recovery}
	\makebox[0.1\textwidth]{\qquad {Object appearance}}
	\makebox[0.3\textwidth]{\qquad\qquad\qquad\quad Surface shape recovery}
	\\
	{\centering\includegraphics[width = 1\linewidth, trim={0px 15px 0px 0px},clip]{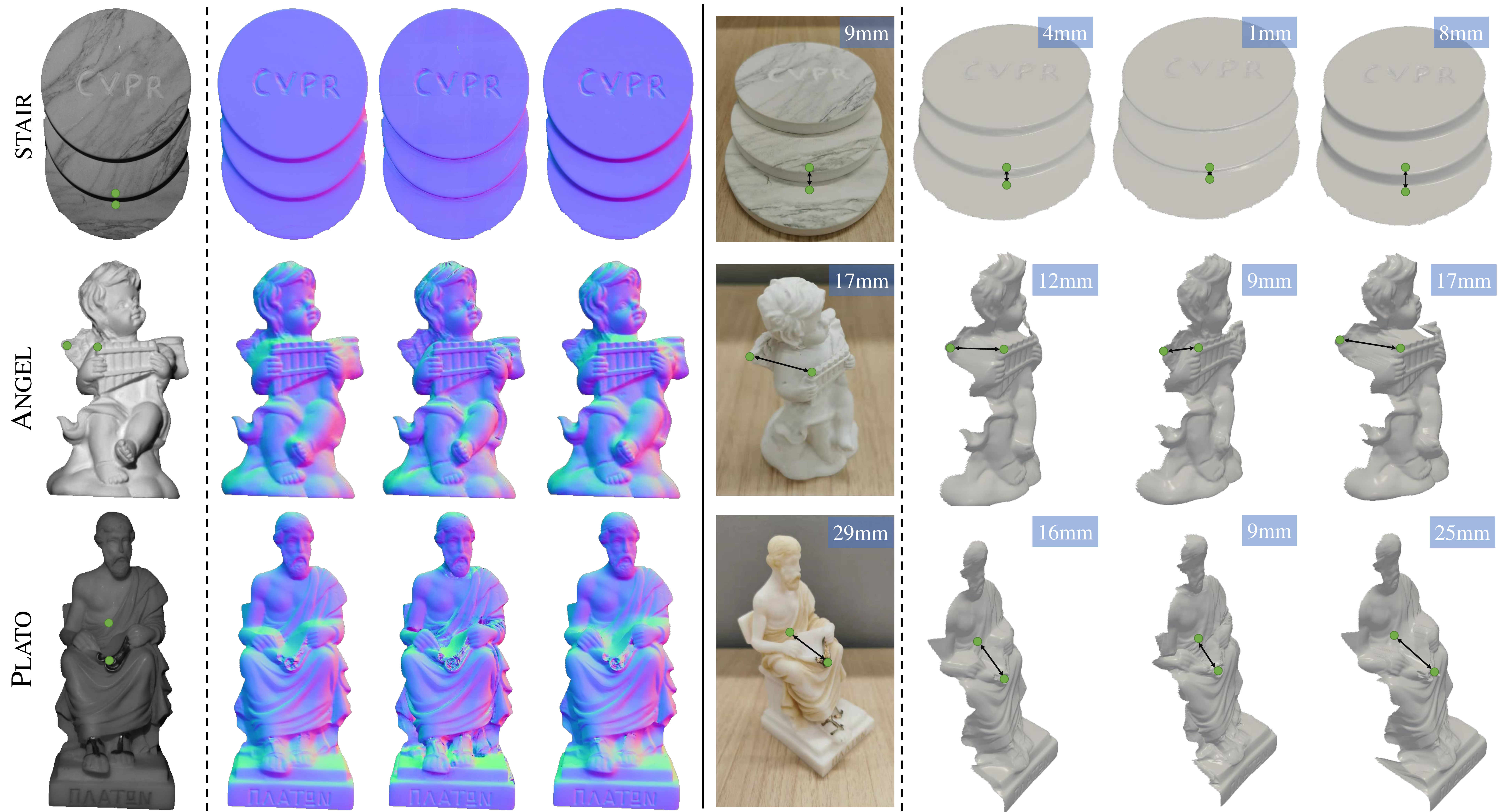}}
	\\
	\makebox[0.08\textwidth]{}
	\makebox[0.11\textwidth]{\qquad\qquad\quad QD18~\cite{queau2018led}}
	\makebox[0.11\textwidth]{\qquad\qquad LB20~\cite{logothetis2020cnn}}
	\makebox[0.12\textwidth]{\qquad\quad Ours}
	\makebox[0.16\textwidth]{}
	\makebox[0.11\textwidth]{\quad QD18~\cite{queau2018led}}
	\makebox[0.11\textwidth]{\qquad\quad LB20~\cite{logothetis2020cnn}}
	\makebox[0.11\textwidth]{\qquad\qquad Ours}
	\caption{Comparison on real data. Values in blue boxes give the measured distance~(object appearance column) and the estimated ones~(shape recovery columns) of two scene points labeled by green dots, revealing the shape estimation accuracy near the discontinuity.}
	\label{fig:real_exp}
		\vspace{-1em}
\end{figure}

\paragraph{Evaluation}
As shown in \fref{fig:real_exp}, we capture three surfaces and compare the surface normal and 3D shape recovery with the baseline methods. For all the objects, the initial depth for QD18~\cite{queau2018led} and LB20~\cite{logothetis2020cnn} are set to $400$~mm, which is close to the true object depth. Based on the visualization of recovered 3D shapes, our method yields more reasonable shape recoveries particularly near depth discontinuities, such as the step area of {\sc Stair}, the wing and neck regions of {\sc Angel}, and the book region of {\sc Plato}. We measure the distance between two scene points of the real surface, as shown in the object appearance column of \fref{fig:real_exp}. The distances between the same two points on our recovered shapes are closer to the measured ones, showing our strength over the baseline methods on shape recovery with edge preservation.
\paragraph{Results on LUCES} We also present a quantitative evaluation of our method on a public near-light photometric stereo dataset LUCES~\cite{mecca2021luces} containing $14$ objects with diverse shapes and materials.
Compared to existing methods on the benchmark~\cite{mecca2021luces}, our method achieved more accurate surface normal and relative depth estimates on $12$ objects in average, except for two metallic surfaces {\sc Bell} and {Cup} that significantly deviate from the Lambertian assumption. More detailed analyses can be found in our supplementary material.

\section{Conclusion}
In this paper, we propose a NLPS method to recover detailed surface shapes with an emphasis on depth edge preservation.
The finite difference approximation used in existing NLPS methods is inaccurate to model the depth derivatives along the depth discontinuities, leading to shape distortions.
Our method avoids the problem by introducing an analytical neural surface representation in NLPS, where depth derivatives can be analytically derived.
Besides, by treating albedos as dependent variables of surface normal and depth under the Lambertian reflectance assumption, we cast NLPS as an optimization of the neural surface parameter only, making our method robust against the initial depth. 
Based on the experiments,
our method shows faithful shape estimates with preserving depth discontinuities and is less sensitive to the initial depth guess. 

\paragraph{Limitations} 
This paper focuses on the geometric aspect of NLPS by proposing an edge-preserving NLPS based on neural surface representation. Although the reconstruction loss in our NLPS method is based on the Lambertian assumption to gain robustness against depth initialization, it is certainly preferable to be able to deal with non-Lambertian reflectances. One potential direction for our future study would be incorporating a parametric/neural reflectance model to represent non-Lambertian reflectances, such as Spherical Gaussian bases~\cite{wang2009all} and MLPs applied in recent multi-view inverse rendering techniques~\cite{zhang2021physg,zhang2021nerfactor}. By optimizing both the reflectance and neural surface parameters using the image reconstruction loss, the surface shape and non-Lambertian reflectances may be jointly recovered. 

\clearpage
%
%
\bibliographystyle{splncs04}
\bibliography{heng}

\begin{thebibliography}{10}
\providecommand{\url}[1]{\texttt{#1}}
\providecommand{\urlprefix}{URL }
\providecommand{\doi}[1]{https://doi.org/#1}

\bibitem{ahmad2014improved}
Ahmad, J., Sun, J., Smith, L., Smith, M.: An improved photometric stereo
  through distance estimation and light vector optimization from diffused
  maxima region. Pattern Recognition Letters  \textbf{50},  15--22 (2014)

\bibitem{bony2013tridimensional}
Bony, A., Bringier, B., Khoudeir, M.: Tridimensional reconstruction by
  photometric stereo with near spot light sources. In: European Signal
  Processing Conference (EUSIPCO). pp.~1--5. IEEE (2013)

\bibitem{burley2012physically}
Burley, B., Studios, W.D.A.: Physically-based shading at {Disney}. In: {Proc.
  of SIGGRAPH} (2012)

\bibitem{clark1992active}
Clark, J.J.: Active photometric stereo. In: {Proc. of IEEE Conference on
  Computer Vision and Pattern Recognition (CVPR)}. pp. 29--34. IEEE (1992)

\bibitem{collins20123d}
Collins, T., Bartoli, A.: {3D} reconstruction in laparoscopy with close-range
  photometric stereo. In: International Conference on Medical Image Computing
  and Computer-Assisted Intervention. pp. 634--642. Springer (2012)

\bibitem{blendercite}
Community, B.O.: Blender - a 3D modelling and rendering package. Blender
  Foundation, Stichting Blender Foundation, Amsterdam (2021),
  \url{http://www.blender.org}

\bibitem{frankot1988method}
Frankot, R.T., Chellappa, R.: A method for enforcing integrability in shape
  from shading algorithms. {IEEE Transactions on Pattern Analysis and Machine
  Intelligence}  \textbf{10}(4),  439--451 (1988)

\bibitem{genova2020local}
Genova, K., Cole, F., Sud, A., Sarna, A., Funkhouser, T.: Local deep implicit
  functions for {3D} shape. In: {Proc. of IEEE Conference on Computer Vision
  and Pattern Recognition (CVPR)}. pp. 4857--4866 (2020)

\bibitem{huang2015near}
Huang, X., Walton, M., Bearman, G., Cossairt, O.: Near light correction for
  image relighting and {3D} shape recovery. In: Digital Heritage. vol.~1, pp.
  215--222. IEEE (2015)

\bibitem{iwahori1990reconstructing}
Iwahori, Y., Sugie, H., Ishii, N.: Reconstructing shape from shading images
  under point light source illumination. In: Proc. of the International
  Conference on Pattern Recognition (ICPR). vol.~1, pp. 83--87. IEEE (1990)

\bibitem{kingma2014adam}
Kingma, D., Ba, J.: Adam: A method for stochastic optimization. In: {Proc. of
  International Conference on Learning Representations (ICLR)} (2015)

\bibitem{kovesi2005shapelets}
Kovesi, P.: Shapelets correlated with surface normals produce surfaces. In:
  {Proc. of International Conference on Computer Vision (ICCV)}. vol.~2, pp.
  994--1001 (2005)

\bibitem{liu2018near}
Liu, C., Narasimhan, S.G., Dubrawski, A.W.: Near-light photometric stereo using
  circularly placed point light sources. In: {Proc. of IEEE International
  Conference on Computatiosnal Photography (ICCP)}. pp. 1--10. IEEE (2018)

\bibitem{logothetis2020cnn}
Logothetis, F., Budvytis, I., Mecca, R., Cipolla, R.: A {CNN} based approach
  for the near-field photometric stereo problem. arXiv preprint
  arXiv:2009.05792  (2020)

\bibitem{logothetis2017semi}
Logothetis, F., Mecca, R., Cipolla, R.: Semi-calibrated near field photometric
  stereo. In: {Proc. of IEEE Conference on Computer Vision and Pattern
  Recognition (CVPR)}. pp. 941--950 (2017)

\bibitem{mecca2021luces}
Mecca, R., Logothetis, F., Budvytis, I., Cipolla, R.: {LUCES}: A dataset for
  near-field point light source photometric stereo. In: Proc. of the British
  Machine Vision Conference (BMVC) (2021)

\bibitem{mecca2016single}
Mecca, R., Qu{\'e}au, Y., Logothetis, F., Cipolla, R.: A single-lobe
  photometric stereo approach for heterogeneous material. SIAM Journal on
  Imaging Sciences  \textbf{9}(4),  1858--1888 (2016)

\bibitem{mecca2015realistic}
Mecca, R., Rodol{\`a}, E., Cremers, D.: Realistic photometric stereo using
  partial differential irradiance equation ratios. Computers \& Graphics
  \textbf{51},  8--16 (2015)

\bibitem{mecca2014near}
Mecca, R., Wetzler, A., Bruckstein, A.M., Kimmel, R.: Near field photometric
  stereo with point light sources. SIAM Journal on Imaging Sciences
  \textbf{7}(4),  2732--2770 (2014)

\bibitem{nie2016novel}
Nie, Y., Song, Z.: A novel photometric stereo method with nonisotropic point
  light sources. In: Proc. of the International Conference on Pattern
  Recognition (ICPR). pp. 1737--1742. IEEE (2016)

\bibitem{oren1994generalization}
Oren, M., Nayar, S.K.: Generalization of {Lambert}'s reflectance model. In:
  Proc. of Annual conference on Computer Graphics and Interactive Techniques.
  pp. 239--246 (1994)

\bibitem{park2014calibrating}
Park, J., Sinha, S.N., Matsushita, Y., Tai, Y.W., So~Kweon, I.: Calibrating a
  non-isotropic near point light source using a plane. In: {Proc. of IEEE
  Conference on Computer Vision and Pattern Recognition (CVPR)}. pp. 2259--2266
  (2014)

\bibitem{park2019deepsdf}
Park, J.J., Florence, P., Straub, J., Newcombe, R., Lovegrove, S.: Deepsdf:
  Learning continuous signed distance functions for shape representation. In:
  {Proc. of IEEE Conference on Computer Vision and Pattern Recognition (CVPR)}.
  pp. 165--174 (2019)

\bibitem{queau2018led}
Qu{\'e}au, Y., Durix, B., Wu, T., Cremers, D., Lauze, F., Durou, J.D.:
  Led-based photometric stereo: Modeling, calibration and numerical solution.
  Journal of Mathematical Imaging and Vision  \textbf{60}(3),  313--340 (2018)

\bibitem{queau2018normal}
Qu{\'e}au, Y., Durou, J.D., Aujol, J.F.: Normal integration: a survey. Journal
  of Mathematical Imaging and Vision  \textbf{60}(4),  576--593 (2018)

\bibitem{santo2020deep}
Santo, H., Waechter, M., Matsushita, Y.: Deep near-light photometric stereo for
  spatially varying reflectances. In: {Proc. of European Conference on Computer
  Vision (ECCV)}. pp. 137--152. Springer (2020)

\bibitem{shi2019benchmark}
Shi, B., Mo, Z., Wu, Z., Duan, D., Yeung, S.K., Tan, P.: A benchmark dataset
  and evaluation for non-{Lambertian} and uncalibrated photometric stereo.
  {IEEE Transactions on Pattern Analysis and Machine Intelligence}  (2019)

\bibitem{silver1980determining}
Silver, W.M.: Determining shape and reflectance using multiple images. Ph.D.
  thesis, Massachusetts Institute of Technology (1980)

\bibitem{sitzmann2020implicit}
Sitzmann, V., Martel, J., Bergman, A., Lindell, D., Wetzstein, G.: Implicit
  neural representations with periodic activation functions. {Proc. of Annual
  Conference on Neural Information Processing Systems (NeurIPS)}  \textbf{33}
  (2020)

\bibitem{Taniai18}
Taniai, T., Maehara, T.: Neural inverse rendering for general reflectance
  photometric stereo. In: Proc. of the International Conference on Machine
  Learning (ICML) (2018)

\bibitem{wang2009all}
Wang, J., Ren, P., Gong, M., Snyder, J., Guo, B.: All-frequency rendering of
  dynamic, spatially-varying reflectance. In: {Proc. of SIGGRAPH}, pp. 1--10
  (2009)

\bibitem{woodham1980ps}
Woodham, R.J.: Photometric method for determining surface orientation from
  multiple images. Optical engineering  (1980)

\bibitem{xie2014surface}
Xie, W., Zhang, Y., Wang, C.C., Chung, R.C.K.: Surface-from-gradients: An
  approach based on discrete geometry processing. In: {Proc. of IEEE Conference
  on Computer Vision and Pattern Recognition (CVPR)}. pp. 2195--2202 (2014)

\bibitem{xu2019disn}
Xu, Q., Wang, W., Ceylan, D., Mech, R., Neumann, U.: Disn: Deep implicit
  surface network for high-quality single-view {3D} reconstruction. arXiv
  preprint arXiv:1905.10711  (2019)

\bibitem{yeh2016streamlined}
Yeh, C.K., Matsuda, N., Huang, X., Li, F., Walton, M., Cossairt, O.: A
  streamlined photometric stereo framework for cultural heritage. In: {Proc. of
  European Conference on Computer Vision (ECCV)}. pp. 738--752. Springer (2016)

\bibitem{zhang2021physg}
Zhang, K., Luan, F., Wang, Q., Bala, K., Snavely, N.: Physg: Inverse rendering
  with spherical gaussians for physics-based material editing and relighting.
  In: {Proc. of IEEE Conference on Computer Vision and Pattern Recognition
  (CVPR)}. pp. 5453--5462 (2021)

\bibitem{zhang2021nerfactor}
Zhang, X., Srinivasan, P.P., Deng, B., Debevec, P., Freeman, W.T., Barron,
  J.T.: Nerfactor: Neural factorization of shape and reflectance under an
  unknown illumination. {ACM Trans. on Graph.}  \textbf{40}(6),  1--18 (2021)

\end{thebibliography}
\end{document}